\documentclass[10pt,twocolumn,letterpaper]{article}

\usepackage{cvpr}
\usepackage{times}
\usepackage{epsfig}
\usepackage{graphicx}
\usepackage{amsmath}
\usepackage{amssymb}

\usepackage[numbers,sort]{natbib}
\usepackage{xcolor, colortbl}
\usepackage{mathtools}
\usepackage{tabularx}
\usepackage{multirow}
\usepackage{enumitem}
\usepackage{bbm}
\usepackage{subfiles}
\usepackage{afterpage}

\newcolumntype{C}[1]{>{\centering\let\newline\\\arraybackslash\hspace{0pt}}p{#1}}

\definecolor{brown}{rgb}{0.80, 0.16, 0.16}
\newcommand{\suha}[1]{{\color{brown}{#1}}}

\definecolor{purp}{rgb}{0.65, 0.16, 0.65}

\definecolor{grey}{rgb}{0.9, 0.9, 0.9}
\newcommand{\ccol}{\cellcolor{grey}}

\usepackage[pagebackref=true,breaklinks=true,letterpaper=true,colorlinks,bookmarks=false]{hyperref}

\cvprfinalcopy 

\def\cvprPaperID{2584} 

\begin{document}

\title{Proxy Anchor Loss for Deep Metric Learning}


\author{
Sungyeon Kim \qquad
Dongwon Kim \qquad
Minsu Cho \qquad
Suha Kwak\\
POSTECH, Pohang, Korea\\
{\tt\small \{tjddus9597, kdwon, mscho, suha.kwak\}@postech.ac.kr}
}

\maketitle
\thispagestyle{empty}


\begin{abstract}
\label{sec:abstract}
Existing metric learning losses can be categorized into two classes: pair-based and proxy-based losses. 
The former class can leverage fine-grained semantic relations between data points, but slows convergence in general due to its high training complexity.
In contrast, the latter class enables fast and reliable convergence, but cannot consider the rich data-to-data relations.
This paper presents a new proxy-based loss that takes advantages of both pair- and proxy-based methods and overcomes their limitations.
Thanks to the use of proxies, our loss boosts the speed of convergence and is robust against noisy labels and outliers.
At the same time, it allows embedding vectors of data to interact with each other through its gradients to exploit data-to-data relations.
Our method is evaluated on four public benchmarks, where a standard network trained with our loss achieves state-of-the-art performance and most quickly converges.
\end{abstract}


\section{Introduction}
\label{sec:intro}

Learning a semantic distance metric has been a crucial step for many applications such as content-based image retrieval~\cite{movshovitz2017no,songCVPR16,Sohn_nips2016,kim2019deep}, face verification~\cite{Schroff2015, liu2017sphereface}, person re-identification~\cite{Chen_2017_CVPR,xiao2017joint}, few-shot learning~\cite{snell2017prototypical, sung2018learning, Qiao_2019_ICCV}, and representation learning~\cite{kim2019deep,Wang2015,Zagoruyko_CVPR_2015}.
Following their great success in visual recognition, deep neural networks have been employed recently for metric learning.
The networks are trained to project data onto an embedding space
in which semantically similar data (\eg, images of the same class) are closely grouped together.
Such a quality of the embedding space is given mainly by loss functions used for training the networks, and most of the losses are categorized into two classes: \emph{pair-based} and \emph{proxy-based}.

The pair-based losses are built upon pairwise distances between data in the embedding space.
A seminal example is Contrastive loss~\cite{Chopra2005,Hadsell2006}, which aims to minimize the distance between a pair of data if their class labels are identical and to separate them otherwise.
Recent pair-based losses consider a group of pairwise distances to handle relations between more than two data~\cite{Schroff2015,Wang2014,songCVPR16,Sohn_nips2016,Yu_2019_ICCV,wang2019multi,wang2019ranked,kim2019deep}.
These losses provide rich supervisory signals for training embedding networks by comparing data to data and examining fine-grained relations between them, \ie, \emph{data-to-data relations}.
However, since they take a tuple of data as a unit input, the losses cause prohibitively high training complexity\footnote{The training complexity indicates the amount of computation required to address the entire training dataset~\cite{aziere2019ensemble, Do_2019_CVPR, Harwood_2017_ICCV, Qian_2019_ICCV, wang2019ranked}.}, $O(M^2)$ or $O(M^3)$ where $M$ is the number of training data, thus slow convergence.
Furthermore, some tuples do not contribute to training or even degrade the quality of the learned embedding space.
To resolve these issues, learning with the pair-based losses often entails tuple sampling techniques~\cite{Schroff2015,sampling_matters,Yuan_2017_ICCV,Harwood_2017_ICCV}, which however have to be tuned by hand and may increase the risk of overfitting.

\begin{figure} [!t]
\centering
\includegraphics[width = 1 \columnwidth]{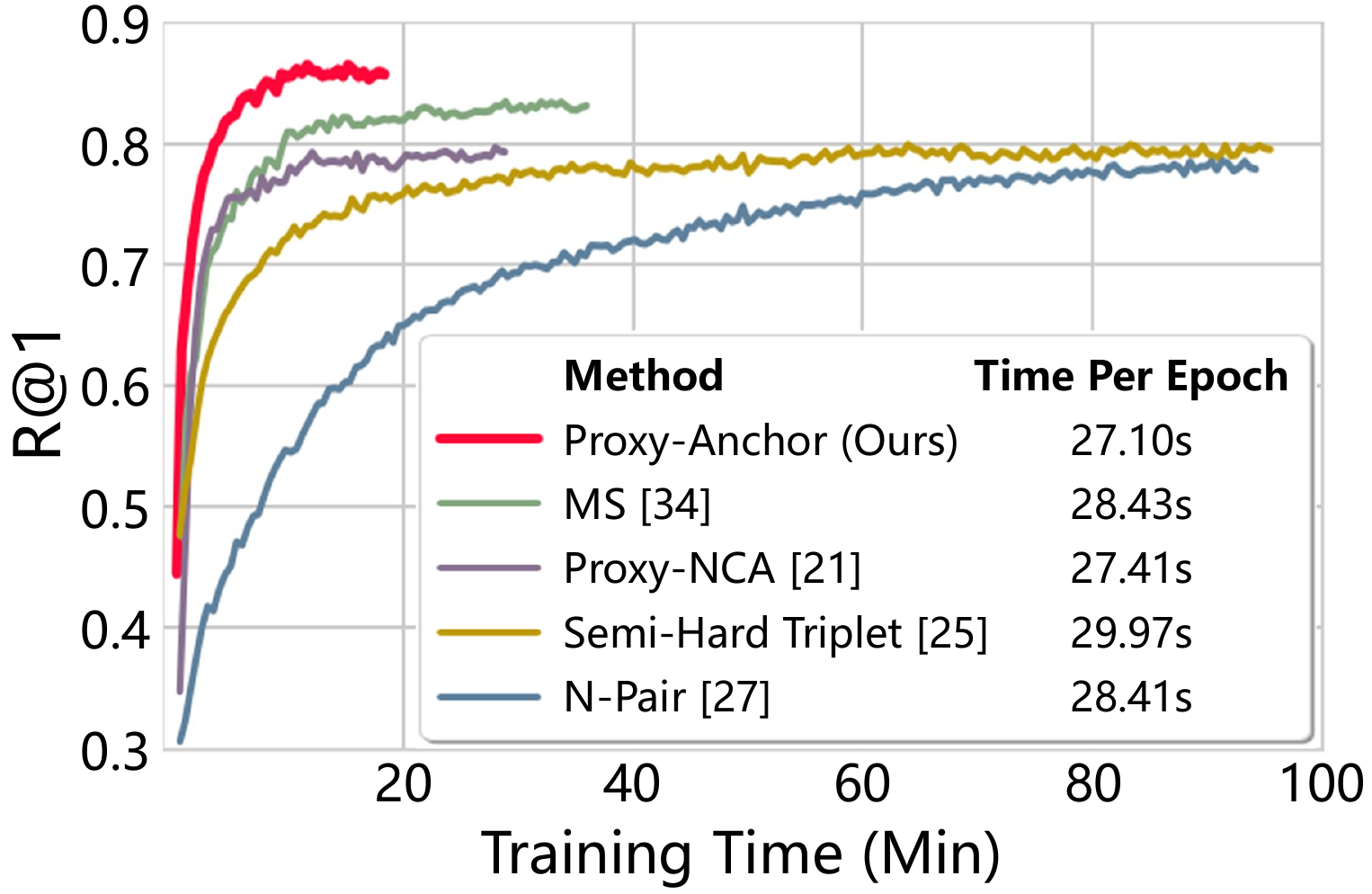}
\caption{
Accuracy in Recall@1 versus training time on the Cars-196~\cite{krause20133d} dataset. Note that all methods were trained with batch size of 150 on a single Titan Xp GPU.
Our loss enables to achieve the highest accuracy, and converge faster than the baselines in terms of both the number of epochs and the actual training time.
} 
\label{fig:convergence}
\vspace{-1mm}
\end{figure}

\begin{figure*} [!t]
\centering
\includegraphics[width = 0.94 \textwidth]{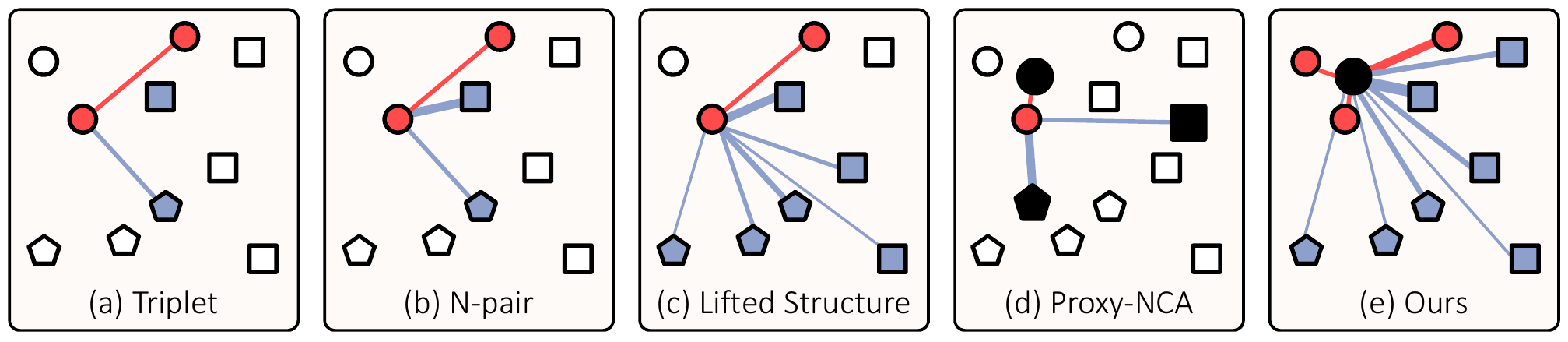}
\caption{
Comparison between popular metric learning losses and ours.
Small nodes are embedding vectors of data in a batch, and 
black ones indicate proxies; their different shapes represent distinct classes.
The associations defined by the losses are expressed by edges, and thicker edges get larger gradients.
Also, embedding vectors associated with the anchor are colored in red if they are of the same class of the anchor (\ie, positive) and in blue otherwise (\ie, negative). 
(a) Triplet loss~\cite{Wang2014, Schroff2015} associates each anchor with a positive and a negative data point without considering their hardness.
(b) $N$-pair loss~\cite{Sohn_nips2016} and (c) Lifted Structure loss~\cite{songCVPR16} reflect hardness of data, but do not utilize all data in the batch.
(d) Proxy-NCA loss~\cite{movshovitz2017no} cannot exploit data-to-data relations since it associates 
each data point only with proxies. 
(e) Our loss handles entire data in the batch, and associates them with each proxy with consideration of their relative hardness determined by data-to-data relations. 
See the text for more details.
} 
\vspace{-2mm}
\label{fig:comparison}
\end{figure*}

The proxy-based losses resolve the above complexity issue by introducing \emph{proxies}~\cite{movshovitz2017no,aziere2019ensemble,Qian_2019_ICCV}.
A proxy is a representative of a subset of training data and learned as a part of the network parameters. 
Existing losses in this category consider each data point as an anchor, associate it with proxies instead of other images, and encourage the anchor to be close to proxies of the same class and far apart from those of different classes. 
Proxy-based losses reduce the training complexity and enable faster convergence since the number of proxies is substantially smaller than that of training data in general.
Further, these losses tend to be more robust against label noises and outliers.
However, since they associate each data point only with proxies, proxy-based losses can leverage only \emph{data-to-proxy relations}, which are impoverished compared to the rich data-to-data relations available for pair-based losses.

In this paper, we propose a novel proxy-based loss called Proxy-Anchor loss, which takes good points of both proxy-based and pair-based losses while correcting their defects. 
Unlike the existing proxy-based losses, the proposed loss utilizes each proxy as an anchor and associates it with all data in a batch.
Specifically, for each proxy, the loss aims to pull data of the same class close to the proxy and to push others away in the embedding space. 
Due to the use of proxies, our loss boosts the speed of convergence with no hyperparameter for tuple sampling, and is robust against noisy labels and outliers.
At the same time, it can take data-to-data relations into account like pair-based losses;
this property is given by associating all data in a batch with each proxy so that the gradients with respect to a data point are weighted by its relative proximity to the proxy (\ie, relative hardness) affected by the other data in the batch.
Thanks to the above advantages, a standard embedding network trained with our loss achieves state-of-the-art accuracy and most quickly converges as shown in Figure~\ref{fig:convergence}.
The contribution of this paper is three-fold: 
\vspace{-2mm}
\begin{itemize}[leftmargin=*]
   \setlength\itemsep{-1mm}
   \item We propose a novel metric learning loss that takes advantages of both pair-based and proxy-based methods; it leverages rich data-to-data relations and enables fast and reliable convergence.
   \item A standard embedding network trained with our loss achieves state-of-the-art performance on the four public benchmarks for metric learning~\cite{CUB200,krause20133d,songCVPR16,DeepFashion}.
   \item Our loss speeds up convergence greatly without careful data sampling; its convergence is even faster than those of  Proxy-NCA~\cite{movshovitz2017no} and Multi-Similarity loss~\cite{wang2019multi}.
\end{itemize}

\section{Related Work}
\label{sec:relatedwork}
In this section, we categorize metric learning losses into two classes, pair-based and proxy-based losses, then review relevant methods for each category. 

\subsection{Pair-based Losses}
\label{sec:pair_loss}

Contrastive loss~\cite{Bromley1994, Chopra2005, Hadsell2006} and Triplet loss~\cite{Wang2014, Schroff2015} are seminal examples of loss functions for deep metric learning.
Contrastive loss takes a pair of embedding vectors as input, and aims to pull them together if they are of the same class and push them apart otherwise.
Triplet loss considers a data point as an anchor, associates it with a positive and a negative data point, and constrains the distance of the anchor-positive pair to be smaller than that of the anchor-negative pair in the embedding space as illustrated in Figure~\ref{fig:comparison}(a).

Recent pair-based losses aim to leverage higher order relations between data and reflect their hardness for further enhancement.
As generalizations of Triplet loss, $N$-pair loss~\cite{Sohn_nips2016} and Lifted Structure loss~\cite{songCVPR16} associate an anchor with a single positive and multiple negative data points, and pull the positive to the anchor and push the negatives away from the anchor while considering their hardness.
As shown in Figure~\ref{fig:comparison}(b)~and~\ref{fig:comparison}(c), however, these losses do not utilize entire data in a batch since they sample the same number of data per negative class, thus may drop informative examples during training.
In contrast, Ranked List loss~\cite{wang2019ranked} takes into account all positive and negative data in a batch and aims to separate the positive and negative sets. 
Multi-Similarity loss~\cite{wang2019multi} also considers every pair of data in a batch, and assigns a weight to each pair according to three complementary types of similarity to focus more on useful pairs 
for improving performance and convergence speed.

Pair-based losses enjoy rich and fine-grained data-to-data relations as they examine tuples (\ie, data pairs or their combinations) during training.
However, since the number of tuples increases polynomially with the number of training data, 
their training complexity is prohibitively high and convergence is slow.
In addition, a large amount of tuples are not effective and sometimes even degrade the quality of the learned embedding space~\cite{Schroff2015,sampling_matters}. 
To address this issue, most pair-based losses entail tuple sampling techniques~\cite{Schroff2015,sampling_matters,Harwood_2017_ICCV,Yuan_2017_ICCV} to select and utilize tuples that will contribute to training.
However, these techniques involve hyperparameters that have to be tuned carefully, and may increase the risk of overfitting since they rely mostly on local pairwise relations within a batch. 
Another way to alleviating the complexity issue is to assign larger weights to more useful pairs during training as in~\cite{wang2019multi}, which however also incorporates a sampling technique.

Our loss resolves this complexity issue by adopting proxies, which enables faster and more reliable convergence compared to pair-based losses.
Furthermore, it demands no additional hyperparameter for tuple sampling.


\subsection{Proxy-based Losses}
\label{sec:proxy_loss}
Proxy-based metric learning is a relatively new approach that can address the complexity issue of the pair-based losses.
A proxy means a representative of a subset of training data and is estimated as a part of the embedding network parameters. 
The common idea of the methods in this category is to infer a small set of proxies that capture the global structure of an embedding space and relate each data point with the proxies instead of the other data points during training.
Since the number of proxies is significantly smaller than that of training data, the training complexity can be reduced substantially.

The first proxy-based loss is Proxy-NCA~\cite{movshovitz2017no}, which is an approximation of Neighborhood Component Analysis (NCA)~\cite{goldberger2005} using proxies.
In its standard setting, Proxy-NCA loss assigns a single proxy for each class, associates a data point with proxies, and encourages the positive pair to be close and negative pairs to be far apart, as illustrated in Figure~\ref{fig:comparison}(d).
SoftTriple loss~\cite{Qian_2019_ICCV}, an extension of SoftMax loss for classification, is similar to Proxy-NCA yet assigns multiple proxies to each class to reflect intra-class variance. 
Manifold Proxy loss~\cite{aziere2019ensemble} is an extension of $N$-pair loss using proxies, and improves the performance by adopting a manifold-aware distance instead of Euclidean distance to measure the semantic distance in the embedding space.

Using proxies in these losses helps improve training convergence greatly, but has an inherent limitation as a side effect: Since each data point is associated only with proxies, the rich data-to-data relations that are available for the pair-based methods are not accessible anymore. 
Our loss can overcome this limitation since its gradients reflect relative hardness of data and allow their embedding vectors to interact with each other during training.



\section{Our Method}
\label{sec:method}



We propose a new metric learning loss called Proxy-Anchor loss to overcome the inherent limitations of the previous methods.
The loss employs proxies that enable fast and reliable convergence as in proxy-based losses. 
Also, although it is built upon data-proxy relations, our loss can utilize data-to-data relations during training like pair-based losses since it enables embedding vectors of data points to be affected by each other through its gradients.
This property of our loss improves the quality of the learned embedding space substantially.

In this section, we first review Proxy-NCA loss~\cite{movshovitz2017no}, a representative proxy-based loss, for comparison to our Proxy-Anchor loss.
We then describe our Proxy-Anchor loss in detail and analyze its training complexity. 


\subsection{Review of Proxy-NCA Loss}
\label{sec:proxy_nca}
In the standard setting, Proxy-NCA loss~\cite{movshovitz2017no} assigns a proxy to each class so that the number of proxies is the same with that of class labels.
Given an input data point as an anchor, the proxy of the same class of the input is regarded as positive and the other proxies are negative. 
Let $x$ denote the embedding vector of the input, $p^+$ be the positive proxy, and $p^-$ be a negative proxy.
The loss is then given by
%
%
%
%
\begin{align}
    \ell(X) & = \sum_{x\in X} -\log \frac{e^{s(x,p^+)}}
        {\displaystyle\sum\limits_{p^{-} \in P^{-}}{e^{s(x,p^-)}}} \\
        & = \sum_{x\in X} \Big\{ -s(x,p^+) + \underset{p^{-} \in P^{-}}{\mathrm{LSE}} s(x,p^-) \Big\}, \label{eq:Proxy_NCA_reform}
\end{align}
where $X$ is a batch of embedding vectors, $P^{-}$ is the set of negative proxies, and $s(\cdot,\cdot)$ denotes the cosine similarity between two vectors.
In addition, $\mathrm{LSE}$ in Eq.~\eqref{eq:Proxy_NCA_reform} means the Log-Sum-Exp function, a smooth approximation to the max function.
The gradient of Proxy-NCA loss with respect to $s(x,p)$ is given by
\begin{align}
    \frac{\partial \ell(X)}{\partial s(x,p)} =
    \begin{dcases}
    -1, & \textrm{if } p = p^+, \\
    \frac{e^{s(x,p)}}{\displaystyle\sum\limits_{p^-\in P^-}{e^{s(x,p^-)}}}, & \textrm{otherwise.}
    \end{dcases} \label{eq:Proxy_NCA_grad}
\end{align}
Eq.~\eqref{eq:Proxy_NCA_grad} shows that minimizing the loss encourages $x$ and $p^{+}$ to be close to each other, and $x$ and $p^{-}$ to be far away.
In particular, $x$ and $p^{+}$ are pulled together by the constant power, while $x$ and $p^{-}$ closer to each other (\ie, harder negative) are more strongly pushed away.

%

Proxy-NCA loss enables fast convergence thanks to its low training complexity, $O(MC)$ where $M$ is the number of training data and $C$ is that of classes, which is substantially lower than $O(M^2)$ or $O(M^3)$ of pair-based losses since $C\ll M$; refer to Section~\ref{sec:complexity_analysis} for details.
Also, proxies are robust against outliers and noisy labels since they are trained to represent groups of data.
However, since the loss associates each embedding vector only with proxies, it cannot exploit fine-grained data-to-data relations.
This drawback limits the capability of embedding networks trained with Proxy-NCA loss. 


\subsection{Proxy-Anchor Loss}
\label{sec:proxy_anchor}
Our Proxy-Anchor loss is designed to overcome the limitation of Proxy-NCA while keeping the low training complexity.
The main idea is to take each proxy as an anchor and associate it with entire data in a batch, as illustrated in Figure~\ref{fig:comparison}(e), so that the data interact with each other through the proxy anchor during training.
Our loss assigns a proxy for each class following the standard proxy assignment setting of Proxy-NCA, and is formulated as
\begin{align}
    \begin{split}
    \ell(X) = & \frac{1}{|P^+|}\sum_{p \in P^+}
    {\log{\bigg(1+\sum_{x\in X_p^+}{e^{-\alpha(s(x,p) - \delta)}}}\bigg)} \\
    & + \frac{1}{|P|}\sum_{p \in P}
    {\log{\bigg(1+\sum_{x\in X_p^-}{e^{\alpha(s(x,p) + \delta)}}}\bigg)},
    \end{split}
    \label{eq:Proxy_Anchor}
\end{align}
where $\delta>0$ is a margin, $\alpha>0$ is a scaling factor, $P$ indicates the set of all proxies, and $P^+$ denotes the set of positive proxies of data in the batch.
Also, for each proxy $p$, a batch of embedding vectors $X$ is divided into two sets: $X^+_p$, the set of positive embedding vectors of $p$, and $X^-_p=X-X^+_p$.
The proposed loss can be rewritten in an easier-to-interpret form as
\begin{align}
    \begin{split}
    \ell(X) = & \frac{1}{|P^+|} \sum_{p \in P^+}
    \bigg[\mathrm{Softplus}\Big(\underset{x\in X_p^+}{\mathrm{LSE}}{-\alpha(s(x,p) - \delta)}\Big)\bigg] \\
    & + \frac{1}{|P|} \sum_{p \in P}
    \bigg[\mathrm{Softplus}\Big(\underset{x\in X_p^-}{\mathrm{LSE}}{\alpha(s(x,p) + \delta)}\Big)\bigg],
    \end{split}
    \label{eq:Proxy_Anchor_reform}
\end{align}
where $\mathrm{Softplus}(z) = \log{(1+e^{z})}, \forall z\in \mathbb{R}$, and is a smooth approximation of ReLU.

\vspace{2mm}
\noindent \textbf{How it works:}
Regarding Log-Sum-Exp as the max function, it is easy to notice that the loss aims to pull $p$ and its most dissimilar positive example (\ie, hardest positive example) together, and to push $p$ and its most similar negative example (\ie, hardest negative example) apart.
Due to the nature of Log-Sum-Exp, 
the loss in practice pulls and pushes all embedding vectors in the batch, but with different degrees of strength that are determined by their relative hardness. 
This characteristic is demonstrated by the gradient of our loss with respect to $s(x,p)$, which is given by
\begin{align}
    \frac{\partial \ell(X)}{\partial s(x,p)} & = 
        \begin{dcases}
            \frac{1}{|P^+|} \ \frac{-\alpha \ h^+_p(x)}{1+\displaystyle\sum\limits_{x'\in X_p^+}{h^+_p(x')}},
            & \forall x \in X_p^+, \\
            \frac{1}{|P|} \ \frac{\alpha \ h^-_p(x)}{1+\displaystyle\sum\limits_{x'\in X_p^-}{h^-_p(x')}},
            & \forall  x \in X_p^-,
        \end{dcases} \label{eq:Proxy_Anchor_grad_simple}
\end{align}
where $h^+_p(x)=e^{-\alpha(s(x,p) - \delta)}$ and $h^-_p(x)=e^{\alpha(s(x,p) + \delta)}$ are positive and negative hardness metrics for embedding vector $x$ given proxy $p$, respectively; $h^+_p(x)$ is large when the positive embedding vector $x$ is far from $p$, and $h^-_p(x)$ is large when the negative embedding vector $x$ is close to $p$.
The scaling parameter $\alpha$ and margin $\delta$ control the relative hardness of data points, and in consequence, determine how strongly pull or push their embedding vectors.

As shown in the above equations, the gradient for $s(x,p)$ is affected by not only $x$ but also other embedding vectors in the batch;
the gradient becomes larger when $x$ is harder than the others.
In this way, our loss enables embedding vectors in the batch to interact with each other and reflects their relative hardness through the gradients, which helps enhance the quality of the learned embedding space.


\begin{figure*} [!t]
\centering
\includegraphics[width = 1 \textwidth]{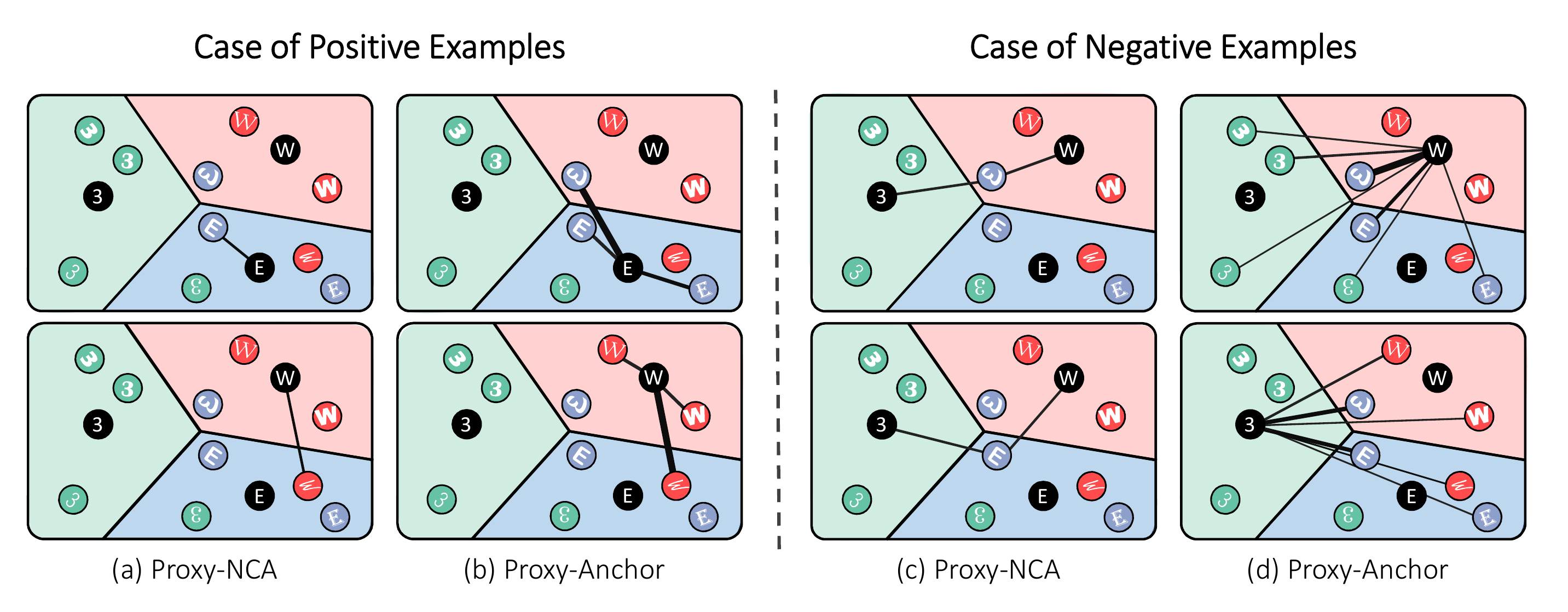}
\caption{
Differences between Proxy-NCA and Proxy-Anchor in handling proxies and embedding vectors during training.
Each 
proxy is colored in black and three different colors indicate distinct classes. 
The associations defined by the losses are expressed by edges, and thicker edges get larger gradients.
(a) Gradients of Proxy-NCA loss with respect to positive examples have the same scale regardless of their hardness.
(b) Proxy-Anchor loss dynamically determines gradient scales regarding relative hardness of all positive examples so as to pull harder positives more strongly.
(c) In Proxy-NCA, each negative example is pushed only by a small number of proxies without considering the distribution of embedding vectors in fine details.
(d) Proxy-Anchor loss considers the distribution of embedding vectors in more details as it has all negative examples affect each other in their gradients.
} 
\label{fig:proxy_losses_comparision}
\end{figure*}

\vspace{2mm}
\noindent \textbf{Comparison to Proxy-NCA:}
The key difference and advantage of Proxy-Anchor over Proxy-NCA is the active consideration of relative hardness based on data-to-data relations.
This property enables Proxy-Anchor loss to provide richer supervisory signals to embedding networks during training.
The gradients of the two losses demonstrate this clearly.
In Proxy-NCA loss, the scale of the gradient is constant for every positive example and that of a negative example is calculated by taking only few proxies into account as shown in Eq.~\eqref{eq:Proxy_NCA_grad}.
In particular, the constant gradient scale for positive examples damages the flexibility and generalizability of embedding networks~\cite{sampling_matters}. 
In contrast, Proxy-Anchor loss determines the scale of the gradient by taking relative hardness into consideration for both positive and negative examples as shown in Eq.~\eqref{eq:Proxy_Anchor_grad_simple}. 
This feature of our loss allows the embedding network to consider data-to-data relations that are ignored in Proxy-NCA and observe much larger area of the embedding space during training than Proxy-NCA.
Figure~\ref{fig:proxy_losses_comparision} illustrates these differences between the two losses in terms of handling the relative hardness of embedding vectors.
In addition, unlike Proxy-Anchor loss, the margin imposed in our loss leads to intra-class compactness and inter-class separability, resulting in a more discriminative embedding space. 

\subsection{Training Complexity Analysis}
\label{sec:complexity_analysis}
Let $M$, $C$, $B$, and $U$ denote the numbers of training samples, classes, batches per epoch, and proxies held by each class, respectively.
$U$ is 1 thus ignored in most of proxy-based losses including ours, but is nontrivial for those managing multiple proxies per class such as SoftTriple loss~\cite{Qian_2019_ICCV}.

Table \ref{tab:training_complexity} compares the training complexity of our loss with those of popular pair- and proxy-based losses. 
The complexity of our loss is $O(MC)$ since it compares every proxy with all positive or all negative examples in a batch. 
More specifically, in Eq.~\eqref{eq:Proxy_Anchor}, the complexity of the first summation is $O(MC)$ and that of the second summation is also $O(MC)$, hence the total training complexity is $O(MC)$.
The complexity of Proxy-NCA~\cite{movshovitz2017no} is also $O(MC)$ since each data point is associated with one positive proxy and $C$$-$$1$ negative proxies as can be seen in Eq.~\eqref{eq:Proxy_NCA_reform}. 
On the other hand, SoftTriple loss~\cite{Qian_2019_ICCV}, a modification of SoftMax using multiple proxies per class, associates each data point with $U$ positive proxies and $U(C-1)$ negative proxies. 
The total training complexity of this loss is thus $O(MCU^2)$. 
In conclusion, the complexity of our loss is the same with or even lower than that of other proxy-based losses.


The training complexity of pair-based losses is higher than that of proxy-based ones.
Since Contrastive loss~\cite{Bromley1994, Chopra2005, Hadsell2006} takes a pair of data as input, its training complexity is $O(M^2)$.
On the other hand, Triplet loss that examines triplets of data has complexity $O(M^3)$, which can be reduced by triplet mining strategies.
For example, semi-hard mining~\cite{Schroff2015} reduces the complexity to $O(M^3/B^2)$ by selecting negative pairs that are located within a neighborhood of anchor but sufficiently far from it. 
Similarly, Smart mining~\cite{Harwood_2017_ICCV} lowers the complexity to $O(M^2)$ by sampling hard triplets using an approximated nearest neighbor index. 
However, even with these techniques, the training complexity of Triplet loss is still high.
Like Triplet loss, $N$-pair loss~\cite{Sohn_nips2016} and Lifted Structure loss~\cite{songCVPR16} that compare each positive pair of data to multiple negative pairs also have complexity $O(M^3)$.
The training complexity of these losses becomes prohibitively high as the number of training data $M$ increases, which slows down the speed of convergence as demonstrated in Figure~\ref{fig:convergence}.


\begin{table}
\centering
\begin{tabularx}{\columnwidth} {c | c | c}
    \hline
    \multicolumn{1}{c|}{Type} & \multicolumn{1}{c|}{Loss} & \multicolumn{1}{c}{Training Complexity} \\ \hline 
    \multirow{3}{*}{Proxy}
    &Proxy-Anchor (Ours) & $O(MC)$ \\
    &Proxy-NCA \cite{movshovitz2017no} & $O(MC)$ \\
    &SoftTriple \cite{Qian_2019_ICCV} & $O(MCU^2)$ \\ \hline
    \multirow{5}{*}{Pair} 
    & Contrastive \cite{Bromley1994, Chopra2005, Hadsell2006} & $O(M^2)$ \\
    &Triplet (Semi-Hard) \cite{Schroff2015} & $O(M^3/B^2)$ \\
    &Triplet (Smart) \cite{Harwood_2017_ICCV} & $O(M^2)$ \\
    &$N$-pair  \cite{Sohn_nips2016} & $O(M^3)$ \\
    &Lifted Structure \cite{songCVPR16} & $O(M^3)$ \\ \hline
\end{tabularx}
\vspace{1mm}
\caption{Comparison of training complexities.}
\label{tab:training_complexity}
\vspace{-2mm}
\end{table}

\begin{table*}[!t]
    \centering
    \begin{tabularx}
    {\textwidth} { | >{\centering\arraybackslash}X | >{\centering\arraybackslash}X >{\centering\arraybackslash}X >{\centering\arraybackslash}X >{\centering\arraybackslash}X >{\centering\arraybackslash}X | >{\centering\arraybackslash}X >{\centering\arraybackslash}X >{\centering\arraybackslash}X >{\centering\arraybackslash}X}
    \hline
    \multicolumn{2}{l|}{\multirow{2}{*}[-2mm]{Recall@$K$}} & \multicolumn{4}{c|}{CUB-200-2011} & \multicolumn{4}{c}{Cars-196}\\ \cline{3-10}
    \multicolumn{2}{l|}{}  & 
        1 & 2 & 4 & 8 &
        1 & 2 & 4 & 8\\ \hline 
    \multicolumn{1}{l|}{Clustering$^{64}$~\cite{songCVPR17}} & \multicolumn{1}{c|}{BN}&
       48.2 & 61.4& 71.8& 81.9&
        58.1 & 70.6& 80.3& 87.8 \\ 
    \multicolumn{1}{l|}{Proxy-NCA$^{64}$~\cite{movshovitz2017no}} & \multicolumn{1}{c|}{BN}&
        49.2 & 61.9& 67.9& 72.4& 
        73.2 & 82.4& 86.4& 87.8 \\ 
    \multicolumn{1}{l|}{Smart Mining$^{64}$~\cite{Harwood_2017_ICCV}} & \multicolumn{1}{c|}{G}&
        49.8 & 62.3& 74.1& 83.3& 
        64.7 & 76.2& 84.2& 90.2 \\ 
    \multicolumn{1}{l|}{MS$^{64}$~\cite{wang2019multi}} & \multicolumn{1}{c|}{BN}&
        57.4 & 69.8& 80.0& 87.8&
        77.3 & 85.3& 90.5& 94.2 \\
    \multicolumn{1}{l|}{SoftTriple$^{64}$~\cite{Qian_2019_ICCV}} & \multicolumn{1}{c|}{BN}&
        \underline{60.1} & \underline{71.9}& \underline{81.2}& \underline{88.5}&
        \underline{78.6} & \underline{86.6}&  \underline{91.8}& \underline{95.4}\\
    \multicolumn{1}{l|}{\ccol Proxy-Anchor$^{64}$} & \multicolumn{1}{c|}{\ccol BN}&
       \ccol \textbf{61.7} & \ccol \textbf{73.0}& \ccol \textbf{81.8}& \ccol \textbf{88.8}&
        \ccol \textbf{78.8} & \ccol \textbf{87.0}& \ccol \textbf{92.2}& \ccol \textbf{95.5}
        \\ \hline
    \multicolumn{1}{l|}{Margin$^{128}$~\cite{sampling_matters}} & \multicolumn{1}{c|}{R50}&
        63.6 & 74.4& 83.1& 90.0&
        79.6 & 86.5& 91.9& 95.1\\ 
    \multicolumn{1}{l|}{HDC$^{384}$~\cite{Yuan_2017_ICCV}} & \multicolumn{1}{c|}{G}&
        53.6 & 65.7& 77.0& 85.6&  
        73.7 & 83.2& 89.5& 93.8\\ 
    \multicolumn{1}{l|}{A-BIER$^{512}$~\cite{opitz2018deep}} & \multicolumn{1}{c|}{G}&
        57.5 & 68.7& 78.3& 86.2& 
        82.0 & 89.0& 93.2& 96.1\\ 
    \multicolumn{1}{l|}{ABE$^{512}$~\cite{ensemble_embedding}} & \multicolumn{1}{c|}{G}&
        60.6 & 71.5& 79.8& 87.4& 
        \underline{85.2} & 90.5 & 94.0 & 96.1\\
    \multicolumn{1}{l|}{HTL$^{512}$~\cite{Ge2018DeepML}} & \multicolumn{1}{c|}{BN}&
        57.1 & 68.8& 78.7& 86.5&
        81.4 & 88.0& 92.7& 95.7\\
    \multicolumn{1}{l|}{RLL-H$^{512}$~\cite{wang2019ranked}} & \multicolumn{1}{c|}{BN}&
        57.4 & 69.7& 79.2& 86.9&
        74.0 & 83.6& 90.1& 94.1\\
    \multicolumn{1}{l|}{MS$^{512}$~\cite{wang2019multi}} & 
    \multicolumn{1}{c|}{BN}&
        \underline{65.7} & \underline{77.0}& \underline{86.3}& \underline{91.2}&
        84.1 & 90.4& 94.0 & 96.5\\ 
    \multicolumn{1}{l|}{SoftTriple$^{512}$~\cite{Qian_2019_ICCV}} & \multicolumn{1}{c|}{BN}&
        65.4 & 76.4& 84.5& 90.4&
        84.5 & \underline{90.7}&  \underline{94.5}& \underline{96.9}\\
    \multicolumn{1}{l|}{\ccol Proxy-Anchor$^{512}$} & \multicolumn{1}{c|}{\ccol BN}&
        \ccol \textbf{68.4} & \ccol \textbf{79.2}& \ccol \textbf{86.8}& \ccol \textbf{91.6}& 
        \ccol \textbf{86.1} & \ccol \textbf{91.7}& \ccol \textbf{95.0}& \ccol \textbf{97.3} \\
        \hline
    \multicolumn{1}{l|}{{$^{\dagger}$Contra+HORDE}$^{512}$ \cite{JACOB_2019_ICCV}} & \multicolumn{1}{c|}{BN}&
        {66.3} & {76.7}& {84.7}& {90.6}& 
        {83.9} & {90.3}& {94.1}& {96.3} \\
    \multicolumn{1}{l|}{\ccol $^{\dagger}$Proxy-Anchor$^{512}$} & \multicolumn{1}{c|}{\ccol BN}&
        \ccol \textbf{71.1} & \ccol \textbf{80.4}& \ccol \textbf{87.4}& \ccol \textbf{92.5}& 
        \ccol \textbf{88.3} & \ccol \textbf{93.1}& \ccol \textbf{95.7}& \ccol \textbf{97.5} \\
        \hline
    \end{tabularx}
    \vspace{0.1mm}
    \caption{
        Recall@$K$ ($\%$) on the CUB-200-2011 and Cars-196 datasets. 
        Superscripts denote embedding sizes and $\dagger$ indicates models using larger input images.
        Backbone networks of the models are denoted by abbreviations: G--GoogleNet~\cite{Googlenet}, BN--Inception with batch normalization~\cite{Batchnorm}, R50--ResNet50~\cite{resnet}.
        }
    \label{tab:eval_cub_cars}
    \vspace*{-4mm}
\end{table*}

\section{Experiments}
\label{sec:experiments}

In this section, our method is evaluated and compared to current state-of-the-art on the four benchmark datasets for deep metric learning~\cite{CUB200,krause20133d,DeepFashion,songCVPR16}.
We also investigate the effect of hyperparameters and embedding dimensionality of our loss to demonstrate its robustness. 

\subsection{Datasets}
We employ CUB-200-2011~\cite{CUB200}, Cars-196~\cite{krause20133d}, Stanford Online Product (SOP)~\cite{songCVPR16} and In-shop Clothes Retrieval (In-Shop)~\cite{DeepFashion} datasets for evaluation. 
For CUB-200-2011, we use 5,864 images of its first 100 classes for training and 5,924 images of the other classes for testing. 
For Cars-196, 8,054 images of its first 98 classes are used for training and 8,131 images of the other classes are kept for testing.
For SOP, we follow the standard dataset split in~\cite{songCVPR16} using 59,551 images of 11,318 classes for training and 60,502 images of the rest classes for testing.
Also for In-Shop, we follow the setting in~\cite{DeepFashion} using 25,882 images of the first 3,997 classes for training and 28,760 images of the other classes for testing; the test set is further partitioned into a query set with 14,218 images of 3,985 classes and a gallery set with 12,612 images of 3,985 classes. 


\subsection{Implementation Details}
\label{sec:implementation_details}

\noindent \textbf{Embedding network:} For a fair comparison to previous work, the Inception network with batch normalization~\cite{Batchnorm} pre-trained for ImageNet classification~\cite{Imagenet} is adopted as our embedding network.
We change the size of its last fully connected layer according to the dimensionality of embedding vectors, and $L_2$-normalize the final output.

\vspace{1mm}
\noindent \textbf{Training:} In every experiment, we employ AdamW optimizer~\cite{adamw}, which has the same update step of Adam~\cite{Adamsolver} yet decays the weight separately. 
Our model is trained for 40 epochs with initial learning rate $10^{-4}$ on the CUB-200-2011 and Cars-196, and for 60 epochs with initial learning rate $6\cdot 10^{-4}$ on the SOP and In-shop.
The learning rate for proxies is scaled up $100$ times for faster convergence. 
Input batches are randomly sampled during training.

\vspace{1mm}
\noindent \textbf{Proxy setting:}
We assign a single proxy for each semantic class following Proxy-NCA~\cite{movshovitz2017no}.
The proxies are initialized using a normal distribution to ensure that they are uniformly distributed on the unit hypersphere.

\vspace{1mm}
\noindent \textbf{Image setting:} Input images are augmented by random cropping and horizontal flipping during training while they are center-cropped in testing. 
The default size of cropped images is 224$\times$224 as in most of previous work, but for comparison to HORDE~\cite{JACOB_2019_ICCV}, we also implement models trained and tested with 256$\times$256 cropped images.

\vspace{1mm}
\noindent \textbf{Hyperparameter setting:} $\alpha$ and $\delta$ in Eq.~\eqref{eq:Proxy_Anchor} is set to 32 and $10^{-1}$, respectively, for all experiments.

\begin{table}
\begin{tabularx}{0.48\textwidth} { >{\centering\arraybackslash}X >{\centering\arraybackslash}X >{\centering\arraybackslash}X >{\centering\arraybackslash}X >{\centering\arraybackslash}X}
    \hline
    Recall@$K$ & \multicolumn{1}{|c}{1} & 10 & 100 & \multicolumn{1}{l}{1000} \\ \hline
    \multicolumn{1}{l}{Clustering$^{64}$~\cite{songCVPR17}} & \multicolumn{1}{|c}{67.0} & 83.7& 93.2& -\\
    \multicolumn{1}{l}{Proxy-NCA$^{64}$~\cite{movshovitz2017no}} &\multicolumn{1}{|c}{73.7} &- &- &- \\
    \multicolumn{1}{l}{MS$^{64}$~\cite{wang2019multi}} & \multicolumn{1}{|c}{74.1} & 87.8& 94.7& \textbf{98.2} \\
    \multicolumn{1}{l}{SoftTriple$^{64}$~\cite{Qian_2019_ICCV}} & \multicolumn{1}{|c}{\underline{76.3}} & \textbf{89.1} & \textbf{95.3}& - \\
    \multicolumn{1}{l}{\ccol Proxy-Anchor$^{64}$} & \multicolumn{1}{|c}{\ccol \textbf{76.5}} & \ccol \underline{89.0}& \ccol \underline{95.1}& \ccol \textbf{98.2} \\ \hline
    \multicolumn{1}{l}{Margin$^{128}$~\cite{sampling_matters}} &\multicolumn{1}{|c}{72.7} & 86.2& 93.8& 98.0 \\
    \multicolumn{1}{l}{HDC$^{384}$~\cite{Yuan_2017_ICCV}} & \multicolumn{1}{|c}{69.5} & 84.4& 92.8& 97.7 \\
    \multicolumn{1}{l}{A-BIER$^{512}$~\cite{opitz2018deep}} &\multicolumn{1}{|c}{74.2} & 86.9& 94.0& 97.8 \\
    \multicolumn{1}{l}{ABE$^{512}$~\cite{ensemble_embedding}} &\multicolumn{1}{|c}{76.3} & 88.4& 94.8& 98.2 \\
    \multicolumn{1}{l}{HTL$^{512}$~\cite{Ge2018DeepML}} & \multicolumn{1}{|c}{74.8} & 88.3& 94.8& 98.4 \\ 
    \multicolumn{1}{l}{RLL-H$^{512}$~\cite{wang2019ranked}} & \multicolumn{1}{|c}{76.1} & 89.1& 95.4& - \\
    \multicolumn{1}{l}{MS$^{512}$~\cite{wang2019multi}} & \multicolumn{1}{|c}{78.2} & \underline{90.5}& \underline{96.0}& \textbf{98.7} \\
    \multicolumn{1}{l}{SoftTriple$^{512}$~\cite{Qian_2019_ICCV}} & \multicolumn{1}{|c}{\underline{78.3}} & 90.3 & 95.9& - \\
    \multicolumn{1}{l}{\ccol Proxy-Anchor$^{512}$} & \multicolumn{1}{|c}{\ccol \textbf{79.1}} & \ccol \textbf{90.8}& \ccol \textbf{96.2}& \ccol \textbf{98.7} \\ \hline
    \multicolumn{1}{l}{{$^{\dagger}$Contra+HORDE}$^{512}$ \cite{JACOB_2019_ICCV}} & \multicolumn{1}{|c}{80.1} & 91.3 & 96.2& \textbf{98.7} \\
    \multicolumn{1}{l}{{\ccol $^{\dagger}$Proxy-Anchor$^{512}$}} & \multicolumn{1}{|c}{\ccol \textbf{80.3}} & \ccol \textbf{91.4} & \ccol \textbf{96.4}& \ccol \textbf{98.7} \\
    \hline
\end{tabularx}
\vspace{0.1mm}
\caption{
    Recall@$K$ ($\%$) on the SOP. 
    Superscripts denote embedding sizes and $\dagger$ indicates models using larger input images.
}
\label{tab:eval_sop}
\vspace{-1mm}
\end{table}
\begin{table}
\begin{tabularx}{0.49\textwidth} { | >{\centering\arraybackslash}X | >{\centering\arraybackslash}X >{\centering\arraybackslash}X >{\centering\arraybackslash}X >{\centering\arraybackslash}X}
    \hline
     \multicolumn{1}{l|}{Recall@$K$} & 1 & 10 & 20 & 40 \\ \hline
     \multicolumn{1}{l|}{HDC$^{384}$~\cite{Yuan_2017_ICCV}} & 62.1 & 84.9& 89.0& 92.3\\
     \multicolumn{1}{l|}{HTL$^{128}$~\cite{Ge2018DeepML}} & 80.9 & 94.3& 95.8 & 97.4\\
     \multicolumn{1}{l|}{MS$^{128}$~\cite{wang2019multi}} & \underline{88.0}& \underline{97.2}& \underline{98.1}& \underline{98.7}\\
     \multicolumn{1}{l|}{\ccol Proxy-Anchor$^{128}$} &  \ccol \textbf{90.8}& \ccol \textbf{97.9}& \ccol \textbf{98.5}& \ccol \textbf{99.0} \\\hline
     \multicolumn{1}{l|}{FashionNet$^{4096}$~\cite{DeepFashion}} & 53.0 & 73.0& 76.0& 79.0\\
     \multicolumn{1}{l|}{A-BIER$^{512}$~\cite{opitz2018deep}} & 83.1 & 95.1& 96.9&  97.8\\
     \multicolumn{1}{l|}{ABE$^{512}$~\cite{ensemble_embedding}} & 87.3& 96.7& 97.9& 98.5\\
     \multicolumn{1}{l|}{MS$^{512}$~\cite{wang2019multi}} & \underline{89.7} & \underline{97.9}& \underline{98.5}& \underline{99.1}\\
     \multicolumn{1}{l|}{\ccol Proxy-Anchor$^{512}$} & \ccol \textbf{91.5} & \ccol \textbf{98.1}& \ccol \textbf{98.8}& \ccol \textbf{99.1}\\ \hline
     \multicolumn{1}{l|}{{$^{\dagger}$Contra+HORDE}$^{512}$ \cite{JACOB_2019_ICCV}} & 90.4 & 97.8& 98.4&  98.9\\ 
     \multicolumn{1}{l|}{{\ccol $^{\dagger}$Proxy-Anchor$^{512}$}} & \ccol \textbf{92.6} & \ccol \textbf{98.3}& \ccol \textbf{98.9}&  \ccol \textbf{99.3}\\ \hline
\end{tabularx}
\vspace{0.1mm}
\caption{
    Recall@$K$ ($\%$) on the In-Shop. 
    Superscripts denote embedding sizes and $\dagger$ indicates models using larger input images.
    }
\label{tab:eval_inshop}
\vspace{-2mm}
\end{table}

\subsection{Comparison to Other Methods}

We demonstrate the superiority of our Proxy-Anchor loss quantitatively by evaluating its image retrieval performance on the four benchmark datasets. 
For a fair comparison to previous work, the accuracy of our model is measured in three different settings: 64/128 embedding dimension with the default image size (224$\times$224), 512 embedding dimension with the default image size, and 512 embedding dimension with the larger image size (256$\times$256).

Results on the CUB-200-2011 and Cars-196 datasets are summarized in Table~\ref{tab:eval_cub_cars}.
Our model outperforms all the previous arts including ensemble methods~\cite{opitz2018deep,ensemble_embedding} in all the three settings.  
In particular, on the challenging CUB-200-2011 dataset, it improves the previous best score by a large margin, 2.7$\%$ in Recall@1.
As reported in Table~\ref{tab:eval_sop}, our model also achieves state-of-the-art performance on the SOP dataset.
It outperforms previous models in all the cases except for Recall@10 and Recall@100 with 64 dimensional embedding, but even in these cases it achieves the second best.
Finally, on the In-Shop dataset, it attains the best scores in all the three settings as shown in~Table~\ref{tab:eval_inshop}. 

For all the datasets, our model with the larger crop size and 512 dimensional embedding achieves the state-of-the-art performance.
Also note that our model with the low embedding dimension often outperforms existing models with the high embedding dimension, which suggests that our loss allows to learn a more compact yet effective embedding space.
Last, but not least, our loss boosts the convergence speed greatly as summarized in Figure~\ref{fig:convergence}.

\begin{figure} [!t]
\centering
\includegraphics[width = 1 \columnwidth]{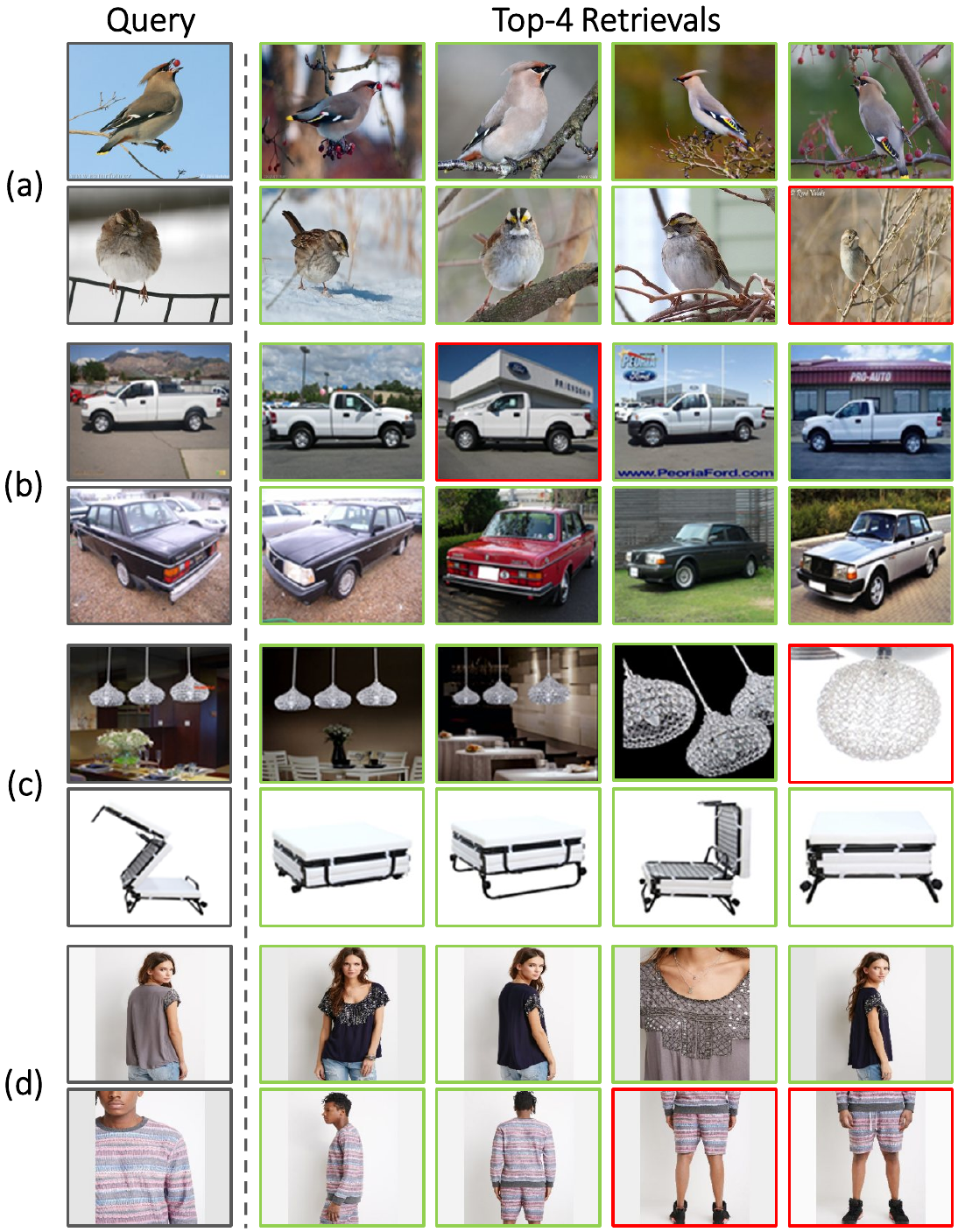}
\caption{Qualitative results on the CUB-200-2011 (a), Cars-196 (b), SOP (c) and In-shop (d). For each query image (\emph{leftmost}), top-4 retrievals are presented. 
The results with red boundary are failure cases, which are however substantially similar to their query images in terms of appearance.
}
\label{fig:qualitative_results}
\end{figure}

\subsection{Qualitative Results}
To further demonstrate the superiority of our loss, we present qualitative retrieval results of our model on the four datasets.
As can be seen in Figure~\ref{fig:qualitative_results}, intra-class appearance variation is significantly large in these datasets in particular by pose variation and background clutter in the CUB200-2011, distinct object colors in the Cars-196, and view-point changes in the SOP and In-Shop datasets.
Even with these challenges, the embedding network trained with our loss performs retrieval robustly.


\subsection{Impact of Hyperparameters}
\noindent \textbf{Batch size:} 
To investigate the effect of batch size on the performance of our loss, we examine Recall@1 of our loss while varying batch size on the four benchmark datasets.
The result of the analysis is summarized in Table~\ref{tab:batch_size_fine} and \ref{tab:batch_size_large}, where one can observe that larger batch sizes improve performance since our loss can consider a larger number of examples and their relations within each batch. 
On the other hand, performance is slightly reduced when the batch size is small since it is difficult to determine the relative hardness in this setting. 
On the datasets with a large number of images and classes, \ie, SOP and In-shop, our loss needs to utilize more examples to fully leverage the relations between data points. 
Our loss achieves the best performance when the batch size is equal to or larger than 300. 


\begin{table}[!t]
\centering
\begin{tabularx}{0.44\textwidth} {>{\centering\arraybackslash}X | >{\centering\arraybackslash}X |>{\centering\arraybackslash}X}
    \hline
    \multicolumn{1}{c|}{\multirow{2}{*}[0mm]{Batch size}} & \multicolumn{2}{c}{Recall@1} \\   \cline{2-3}
    & \multicolumn{1}{c|}{CUB-200-2011} & Cars-196 \\ \hline
    30 & 65.9 & 84.6\\
    60 & 67.0& 86.2\\
    90 & 68.4& 86.2\\
    120 & 68.5& 86.3 \\
    150 & 68.6& \textbf{86.4}\\
    180 & \textbf{69.0}& 86.2\\ \hline
\end{tabularx}
\vspace{1mm}
\caption{Accuracy of our model in Recall@1 versus batch size on the CUB-200-2011 and Cars-196.}
\label{tab:batch_size_fine}
\vspace*{-1mm}
\end{table}
\begin{table}[!t]
\centering
\begin{tabularx}{0.44\textwidth} {>{\centering\arraybackslash}X | >{\centering\arraybackslash}X |>{\centering\arraybackslash}X}
    \hline
    \multicolumn{1}{c|}{\multirow{2}{*}[0mm]{Batch size}} & \multicolumn{2}{c}{Recall@1} \\ \cline{2-3}
    & \multicolumn{1}{c|}{SOP} & In-shop \\ \hline
    30 & 76.0 & 91.3\\
    60 & 78.0& 91.3\\
    90 & 78.5& 91.5\\
    120 & 78.9& 91.7 \\
    150 & 79.1& 91.9\\
    300 & \textbf{79.3}& \textbf{92.0}\\ 
    600 & \textbf{79.3}& 91.7 \\ \hline
\end{tabularx}
\vspace{1mm}
\caption{Accuracy of our model in Recall@1 versus batch size on the SOP and In-shop.}
\label{tab:batch_size_large}
\end{table}

\begin{figure} [!t]
\centering
\includegraphics[width = 0.85 \columnwidth]{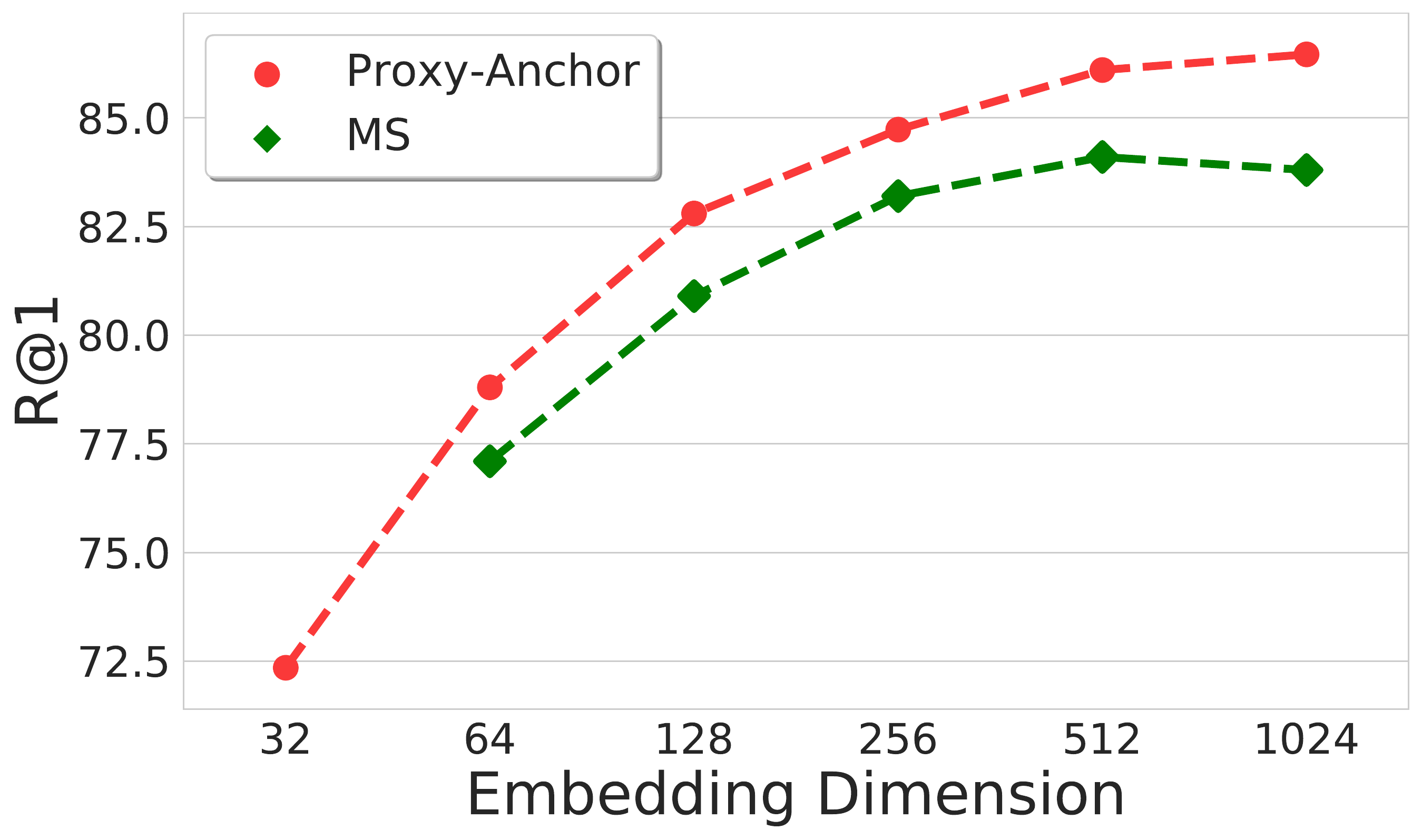}
\caption{Accuracy in Recall@$1$ versus embedding dimension on the Cars-196.
} 
\label{fig:embedding_dim}
\end{figure}

\vspace{1mm}
\noindent \textbf{Embedding dimension:}
The dimension of embedding vectors is a crucial factor that controls the trade-off between speed and accuracy in image retrieval systems.
We thus investigate the effect of embedding dimensions on the retrieval accuracy in our Proxy-Anchor loss.
We test our loss with embedding dimensions varying from 64 to 1,024 following the experiment in~\cite{wang2019multi}, and further examine that with 32 embedding dimension.
The result of analysis is quantified in Figure~\ref{fig:embedding_dim}, in which the retrieval performance of our loss is compared with that of MS loss~\cite{wang2019multi}.
The performance of our loss is fairly stable when the dimension is equal to or larger than 128.
Moreover, our loss outperforms MS loss in all embedding dimensions, and more importantly, its accuracy does not degrade even with the very high dimensional embedding unlike MS loss. 


\begin{figure} [!t]
\centering
\includegraphics[width = 0.88 \columnwidth]{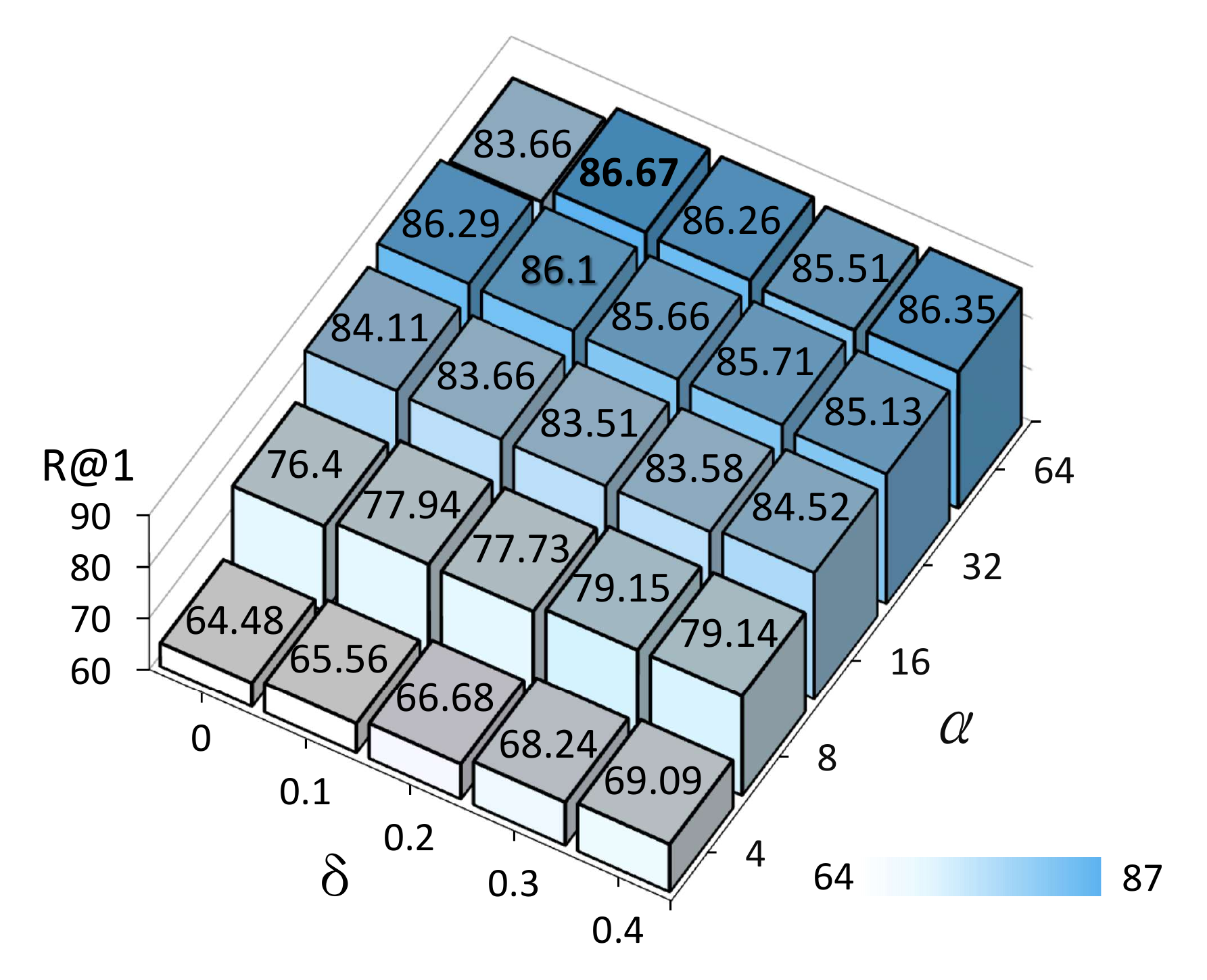}
\caption{Accuracy in Recall@$1$ versus $\delta$ and $\alpha$ on the Cars-196.
} 
\label{fig:hyper_params}
\end{figure}

\vspace{1mm}
\noindent \textbf{$\alpha$ and $\delta$ of our loss:} 
We also investigate the effect of the two hyperparameters $\alpha$ and $\delta$ of our loss on the Cars-196 dataset.
The results of our analysis are summarized in Figure~\ref{fig:hyper_params}, in which we examine Recall@1 of Proxy-Anchor by varying the values of the hyperparameters $\alpha \in \{4, 8, 16, 32, 64\}$ and $\delta \in \{0, 0.1, 0.2, 0.3, 0.4\}$.
The results suggest that when $\alpha$ is greater than 16, the accuracy of our model is high and stable, thus insensitive to the hyperparameter setting.
Our loss outperforms current state-of-the-art with any $\alpha$ greater than 16.
In addition, increasing $\delta$ improves performance although its effect is relatively small when $\alpha$ is large.
Note that our hyperparameter setting reported in Section~\ref{sec:implementation_details} is not the best, although it outperforms all existing methods on the dataset, as we did not tune the hyperparameters to optimize the test accuracy.

\section{Conclusion}
\label{sec:conclusion}
We have proposed a novel metric learning loss that takes advantages of both proxy- and pair-based losses.
Like proxy-based losses, it enables fast and reliable convergence, and like pair-based losses, it can leverage rich data-to-data relations during training.
As a result, our model has achieved state-of-the-art performance on the four public benchmark datasets, and at the same time, converged most quickly with no careful data sampling technique. 
In the future, we will explore extensions of our loss for deep hashing networks to improve its computational efficiency in testing as well as that in training.

\vspace{0.3cm}
{\small
\noindent \textbf{Acknowledgement:} This work was supported by IITP grant, Basic Science Research Program, and R\&D program for Advanced Integrated-intelligence for IDentification through the NRF funded by the Ministry of Science, ICT (No.2019-0-01906 Artificial Intelligence Graduate School Program (POSTECH), NRF-2018R1C1B6001223, NRF-2018R1A5A1060031, NRF-2018M3E3A1057306, NRF-2017R1E1A1A01077999).
}

{\small
\bibliographystyle{ieee_fullname}
\bibliography{cvlab_kwak}
}

\renewcommand\thesection{\Alph{section}}
\setcounter{section}{0}

\newpage
\bigskip~\bigskip~\bigskip~\bigskip~\bigskip~\bigskip~\bigskip~\bigskip~\bigskip~\bigskip~\bigskip~\bigskip~\bigskip~\bigskip~\bigskip~\bigskip~\bigskip~\bigskip~\bigskip~\bigskip~\bigskip~\bigskip~\bigskip~\bigskip~\bigskip~\bigskip~\bigskip~\bigskip~\bigskip~\bigskip~\bigskip~\bigskip~\bigskip~\bigskip~\bigskip~\bigskip~\bigskip~\bigskip~\bigskip~\bigskip~\bigskip~\bigskip~\bigskip~\bigskip~\bigskip~\bigskip~\bigskip~\bigskip~\bigskip~\bigskip~\bigskip~\bigskip~\bigskip~\bigskip~\bigskip~\bigskip~\bigskip~\bigskip~\bigskip~\bigskip~\bigskip~\bigskip~\bigskip

\begin{table*}
    \centering
    \begin{tabularx}
    {\textwidth} {>{\centering\arraybackslash}X | >{\centering\arraybackslash}X | >{\centering\arraybackslash}X >{\centering\arraybackslash}X >{\centering\arraybackslash}X >{\centering\arraybackslash}X | >{\centering\arraybackslash}X >{\centering\arraybackslash}X >{\centering\arraybackslash}X >{\centering\arraybackslash}X}
    \hline
    \multicolumn{1}{c|}{\multirow{2}{*}{Network}} &  \multicolumn{1}{c|}{\multirow{2}{*}{Image Size}}
    & \multicolumn{4}{c|}{CUB-200-2011} & \multicolumn{4}{c}{Cars-196}\\ \cline{3-10}
    & 
        \multicolumn{1}{c|}{\multirow{4}{*}[-4.5mm]{$224 \times 224$}}
        &R@1 & R@2 & R@4 & R@8 &
        R@1 & R@2 & R@4 & R@8\\ \hline 
    \multicolumn{1}{c|}{GoogleNet} &
        &63.8& 74.4& 83.6& 90.4 &
        84.3& 90.4& 94.1& 96.7 \\ 
    \multicolumn{1}{c|}{Inception-BN} & 
        &68.4& 79.2& 86.8& 91.6&
        86.1& 91.7& 95.0& 97.3\\ 
    \multicolumn{1}{c|}{ResNet-50} &
        &69.7& 80.0& 87.0& 92.4&
        87.7& 92.9& 95.8& 97.9 \\
    \multicolumn{1}{c|}{ResNet-101} &
        &\textbf{70.8}& \textbf{81.0}& \textbf{88.1}& \textbf{93.0}&
        \textbf{87.9}& \textbf{93.0}& \textbf{96.1}& \textbf{97.9} \\ \hline
    \multicolumn{1}{c|}{\multirow{3}{*}{Inception-BN}} & 
    \multicolumn{1}{c|}{$256 \times 256$}
        &71.1 & 80.4& 87.4& 92.5&
        88.3 & 93.1& 95.7& 97.5 \\
    & \multicolumn{1}{c|}{$324 \times 324$} 
        &74.0 & 82.9& 88.9& 93.2&
        91.1 & 94.9& 96.9& 98.3 \\
    & \multicolumn{1}{c|}{$448 \times 448$} 
    &\textbf{77.3} & \textbf{85.6}& \textbf{91.1}& \textbf{94.2}&
        \textbf{92.9} & \textbf{96.1}& \textbf{97.7}& \textbf{98.7} \\\hline
    \end{tabularx}
    \vspace{0.1mm}
    \caption{
        Comparison both different backbone networks and different sizes of images on the CUB-200-2011 and Cars-196 datasets.
        }
    \label{tab:eval_cub_cars_supp}
\end{table*}


\section{Appendix}
This appendix presents additional experimental results omitted from the main paper due to the space limit.
Section~\ref{sec:netowrk_imagesize} analyzes the impact of the backbone networks and the size of input images in our framework.
Finally, Section~\ref{sec:qualitative_results} provides t-SNE visualization of the learned embedding space and more qualitative results of image retrieval on the four benchmark datasets~\cite{CUB200,krause20133d,songCVPR16,DeepFashion}.


\subsection{Impact of Backbone Network \& Image Size}
\label{sec:netowrk_imagesize}
Existing methods in deep metric learning have adopted various kinds of backbone networks. In this section, we compare the performance of our loss using popular network architectures as backbone networks on the CUB-200-2011~\cite{CUB200} and Cars-196~\cite{krause20133d}. 
For all experiments, we use 512 dimensional embedding and fix hyperparameters $\alpha$ and $\delta$ to 32 and $10^{-1}$, respectively. In addition to the Inception with batch normalization (Inception-BN) used in the main paper, 
we adopt GoogleNet, ResNet-50, and ResNet-101 as embedding networks trained with our loss. 
The results are summarized in Table~\ref{tab:eval_cub_cars_supp}, 
where a more powerful architecture achieves a better score in general. 
Note that our method with GoogleNet backbone outperforms existing models based on the same backbone, except for ABE~\cite{ensemble_embedding}, an ensemble model, on the Cars-196 dataset.
Furthermore, when using ResNet-50 and ResNet-101 as backbone networks, our model outperforms all the previous methods by large margins.

The main paper showed that the large image size contributed significantly to performance improvement, and we further investigate the performance at larger image size settings. 
We evaluate our method with Inception-BN backbone while varying the sizes of input images: $\{224 \times 224, 256 \times 256, 324 \times 324, 448 \times 448\}$. 
Table~\ref{tab:eval_cub_cars_supp} also shows that the accuracy improves consistently as the sizes of the input images increase. Even our model with images size of $448 \times 448$ has 8.9\% improvement over the default images size ($224 \times 224$) in Recall@1. Increasing the image size decreases the allowable batch size, but with enough GPU memory, using a larger image size is the most effective way to improve performance than using a powerful architecture.

\subsection{Additional Qualitative Results}
\label{sec:qualitative_results}
\begin{figure*} [!t]
\centering
\includegraphics[width = 1 \textwidth]{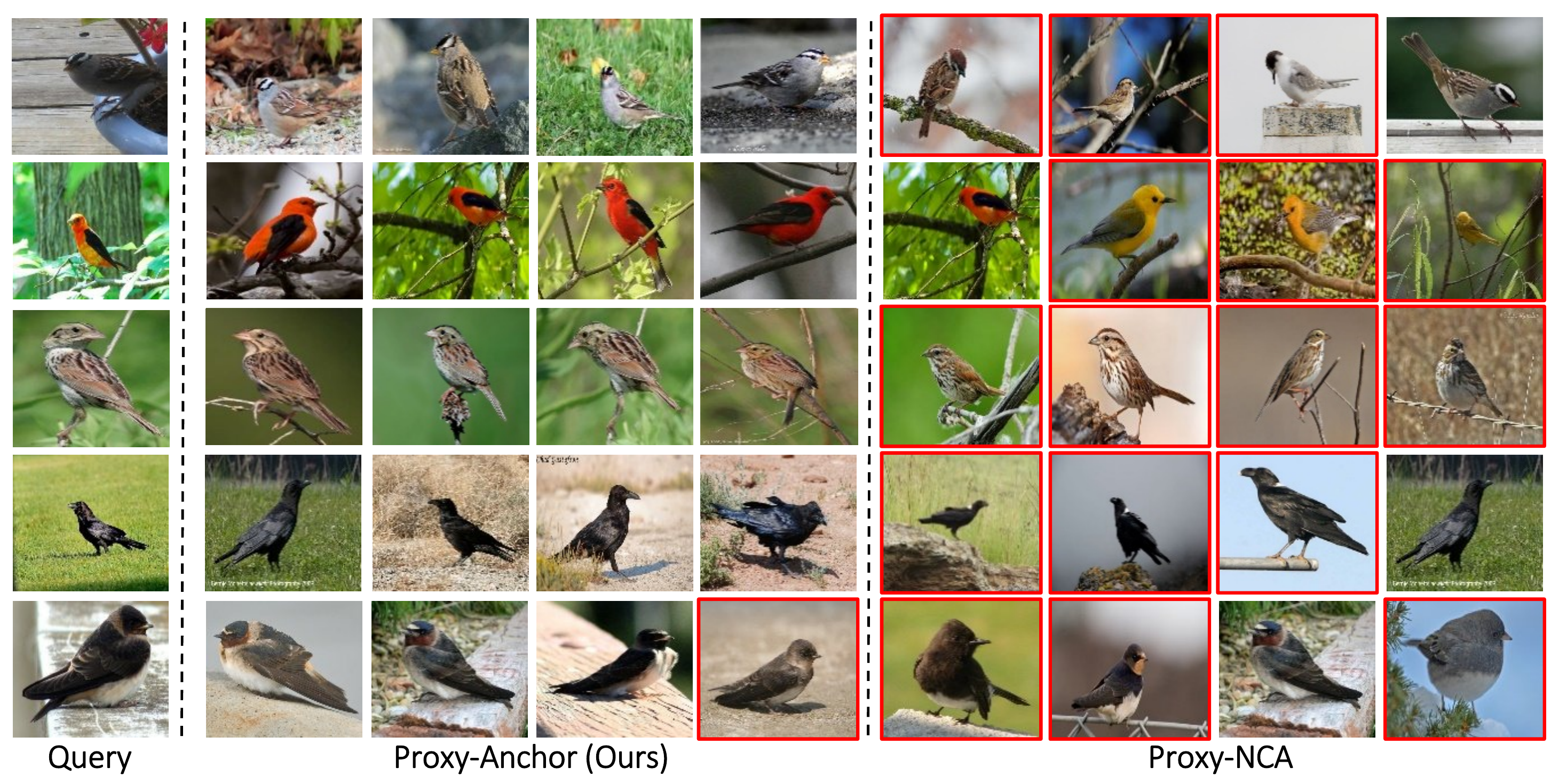}
\caption{Qualitative results on the CUB-200-2011 comparing with Proxy-NCA loss. For each query image (\emph{leftmost}), top-4 retrievals are presented. The result with red boundary is a failure case.
} 
\label{fig:cub_qual}
\end{figure*}

\begin{figure*} [!t]
\centering
\includegraphics[width = 1 \textwidth]{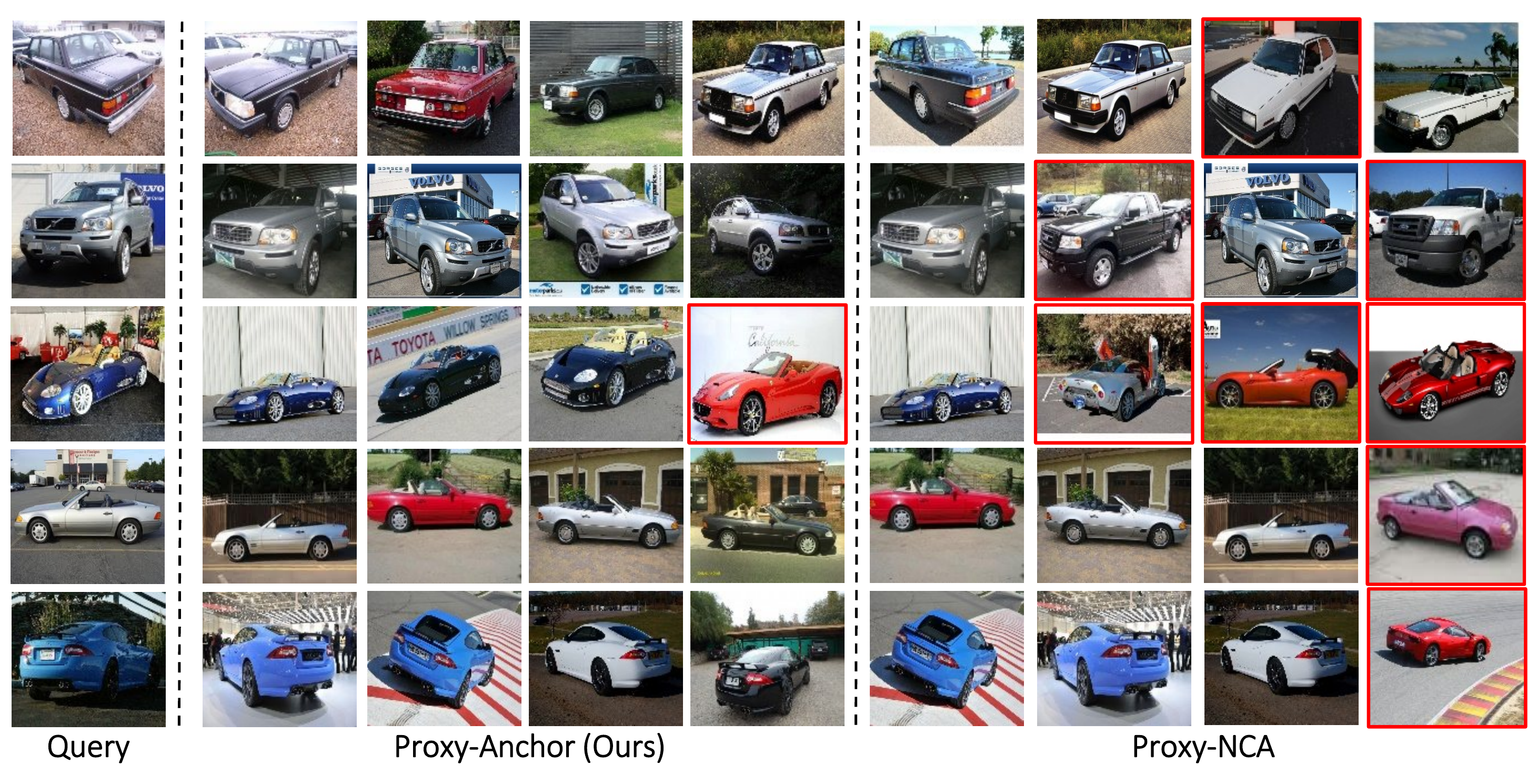}
\caption{Qualitative results on the Cars-196 comparing with Proxy-NCA loss. For each query image (\emph{leftmost}), top-4 retrievals are presented. The result with red boundary is a failure case.
} 
\label{fig:cars_qual}
\end{figure*}

\begin{figure*} [!t]
\centering
\includegraphics[width = 1 \textwidth]{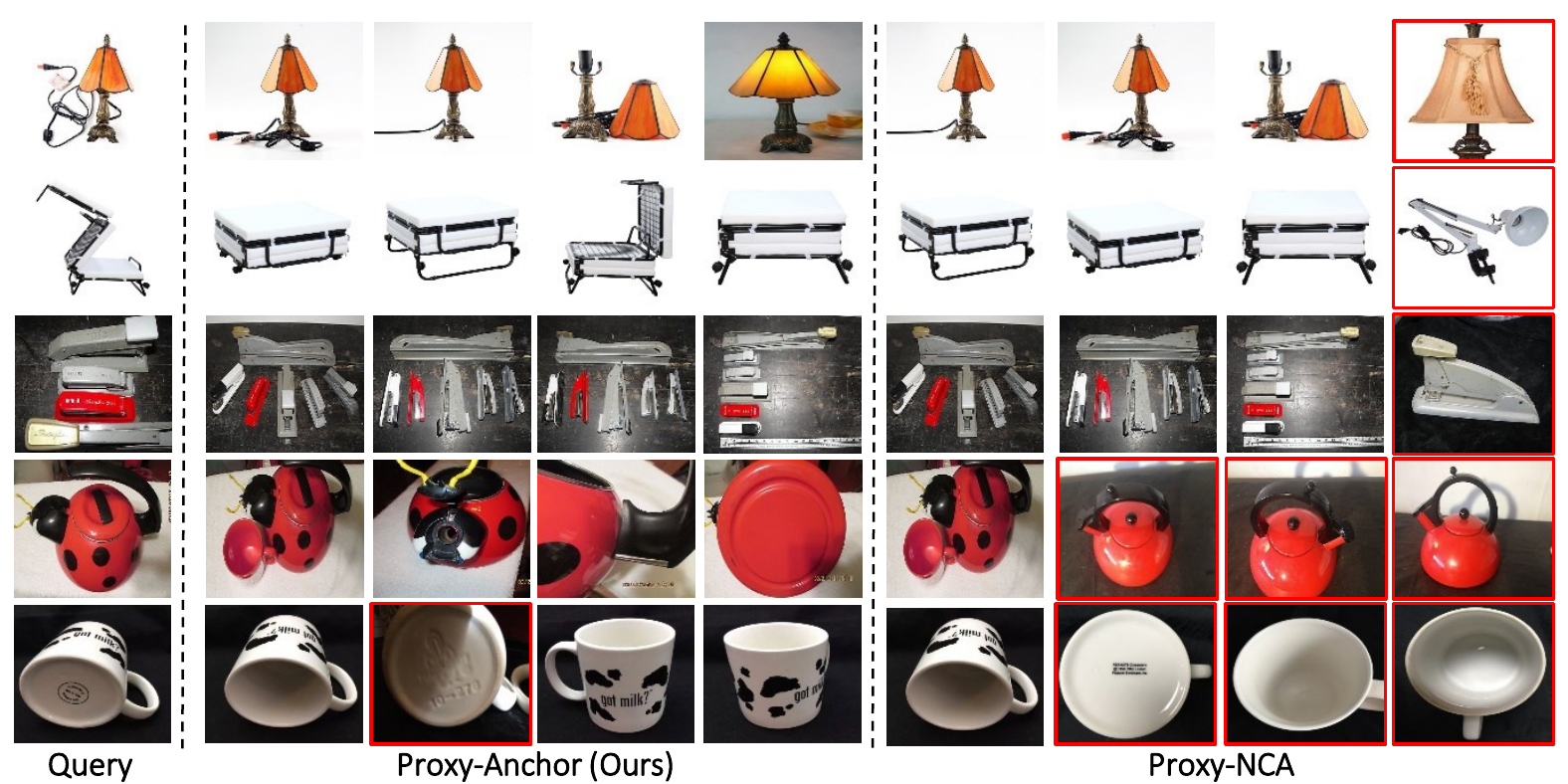}
\caption{Qualitative results on the SOP comparing with Proxy-NCA loss. For each query image (\emph{leftmost}), top-4 retrievals are presented. The result with red boundary is a failure case.
} 
\label{fig:SOP_qual}
\end{figure*}

\begin{figure*} [!t]
\centering
\includegraphics[width = 1 \textwidth]{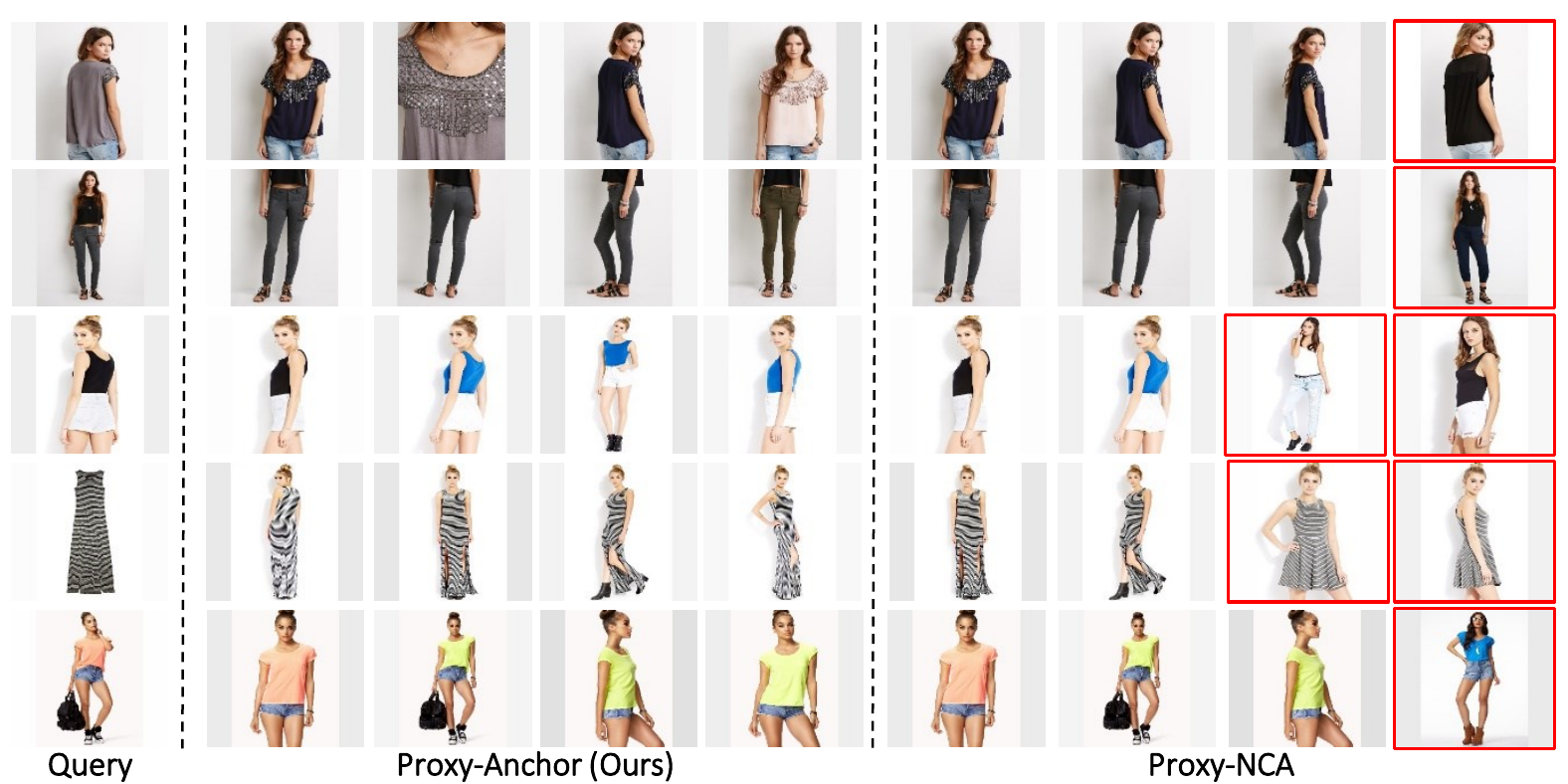}
\caption{Qualitative results on the In-shop comparing with Proxy-NCA loss. For each query image (\emph{leftmost}), top-4 retrievals are presented. The result with red boundary is a failure case.
} 
\label{fig:inshop_qual}
\end{figure*}

More qualitative examples for image retrieval on the CUB-200-2011 and Cars-196 are shown in Figure~\ref{fig:cub_qual} and Figure~\ref{fig:cars_qual}, respectively. The results of our model are compared with those of model trained with Proxy-NCA loss~\cite{movshovitz2017no} using the same backbone network. The overall results indicate that our model learned a higher quality embedding space than the baseline. In the examples in the 3rd and 4th rows of Figure~\ref{fig:cub_qual}, both models retrieved birds with a similar appearance to the query, but only our model produces accurate results. Also, the example in the first row of Figure~\ref{fig:cars_qual} shows successful retrievals despite different view-point changes and colors. Figures~\ref{fig:SOP_qual} and \ref{fig:inshop_qual} compare the qualitative results of the SOP and In-shop datasets. As shown in the 2nd, 4th, and 5th rows of Figure~\ref{fig:SOP_qual}, our model successfully retrieved the same object even with extreme view-point changes. Also, in the 4th row of Figure \ref{fig:inshop_qual}, the baseline is confused with a short dress of similar pattern, whereas our model retrieves the long dress exactly.

\begin{figure*} [!t]
\centering
\includegraphics[width = 1 \textwidth]{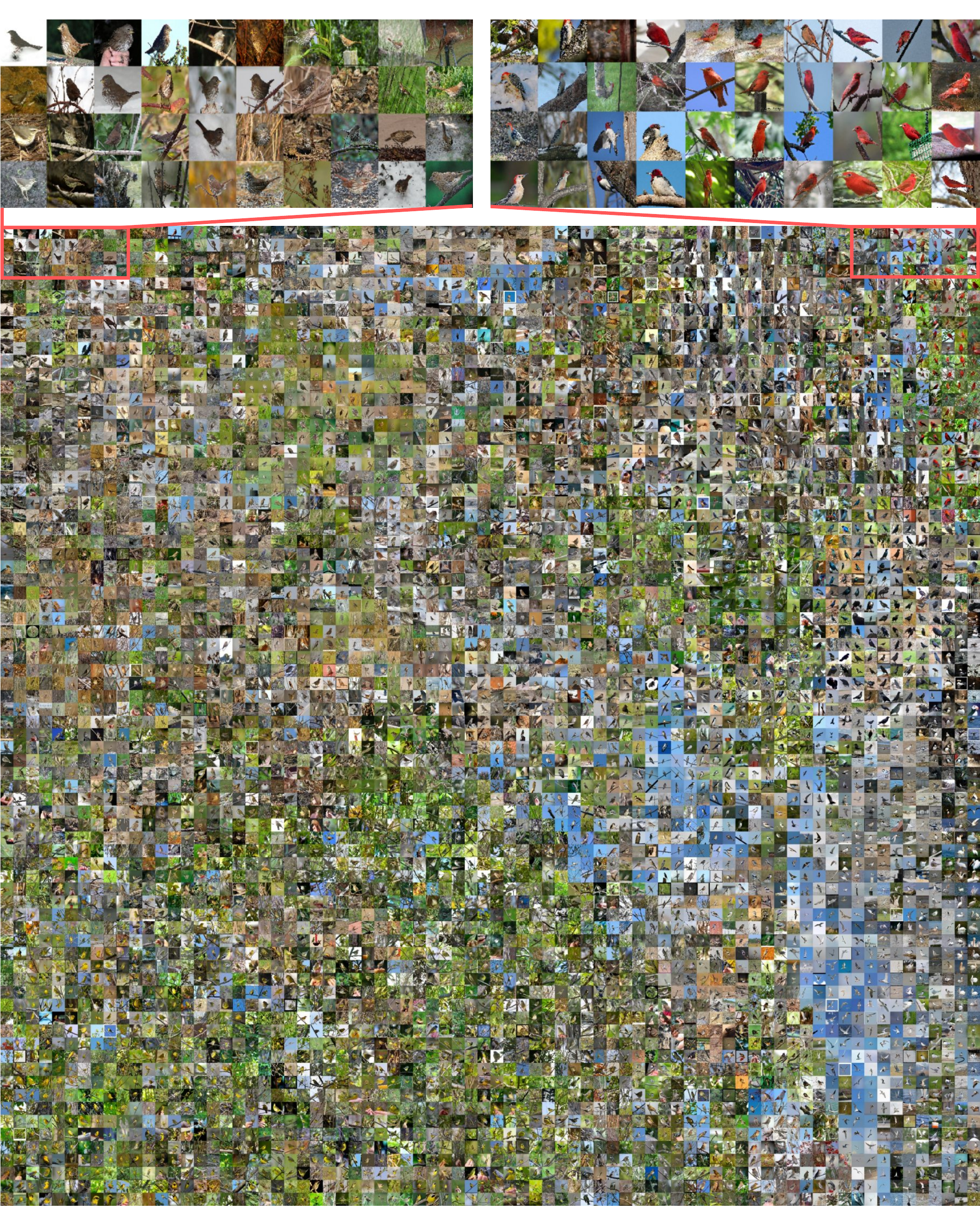}
\caption{t-SNE visualization of our embedding space learned on the test split of CUB-200-2011 dataset in a grid.} 
\label{fig:cub_tsne}
\end{figure*}

\begin{figure*} [!t]
\centering
\includegraphics[width = 1 \textwidth]{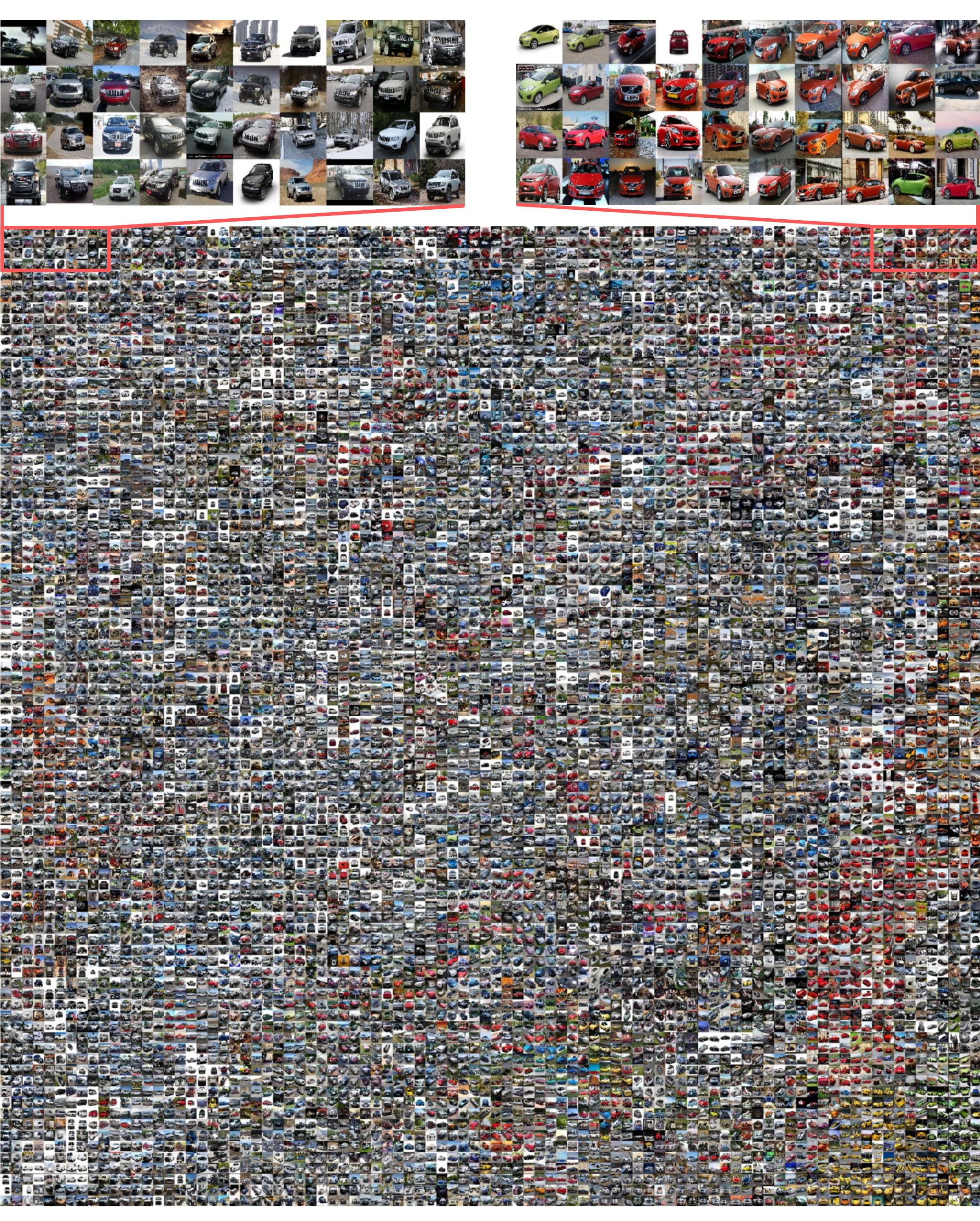}
\caption{t-SNE visualization of our embedding space learned on the test split of Cars-196 dataset in a grid.
} 
\label{fig:cars_tsne}
\end{figure*}

\begin{figure*} [!t]
\centering
\includegraphics[width = 1 \textwidth]{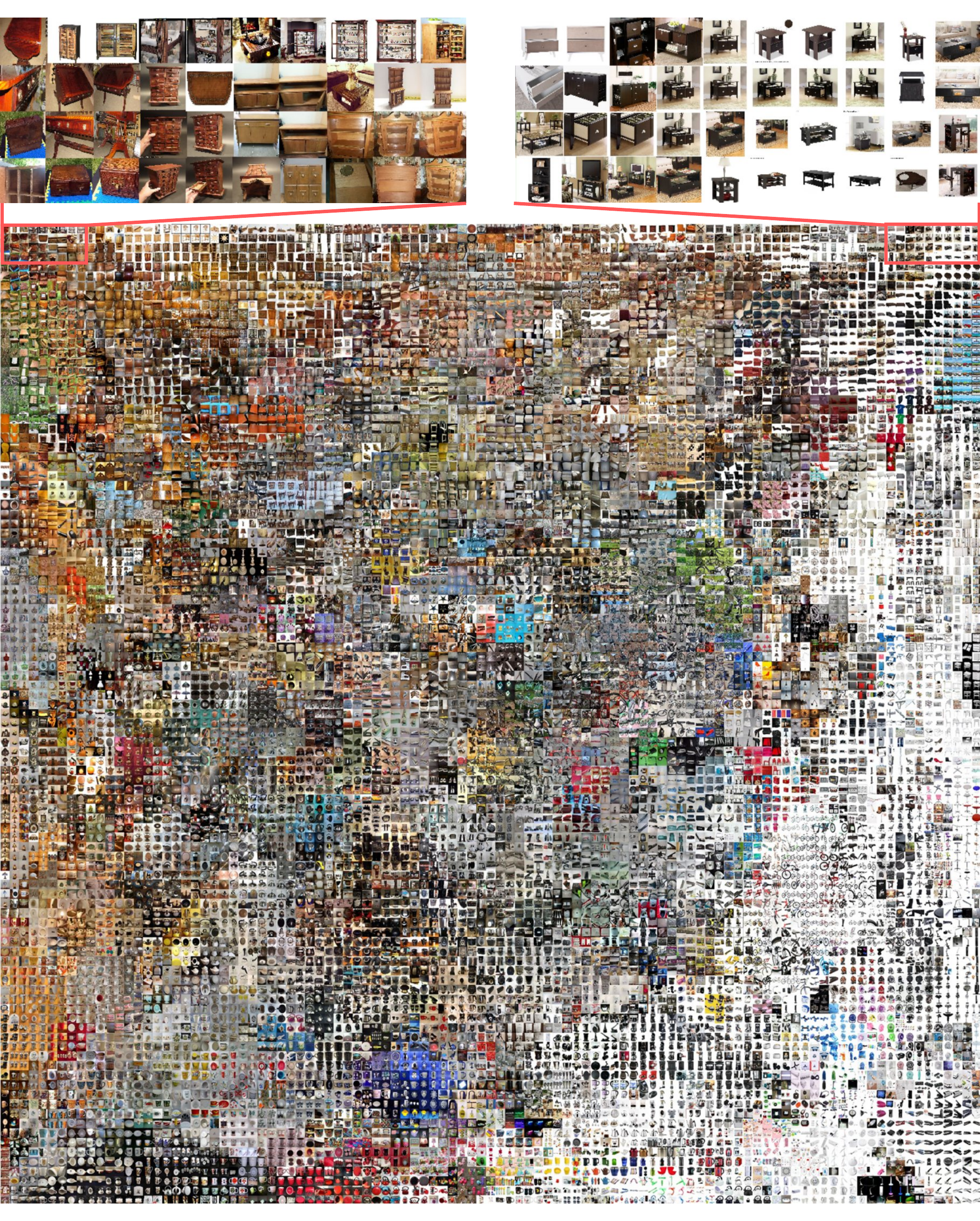}
\caption{t-SNE visualization of our embedding space learned on the test split of SOP dataset in a grid.
} 
\label{fig:SOP_tsne}
\end{figure*}

\begin{figure*} [!t]
\centering
\includegraphics[width = 1 \textwidth]{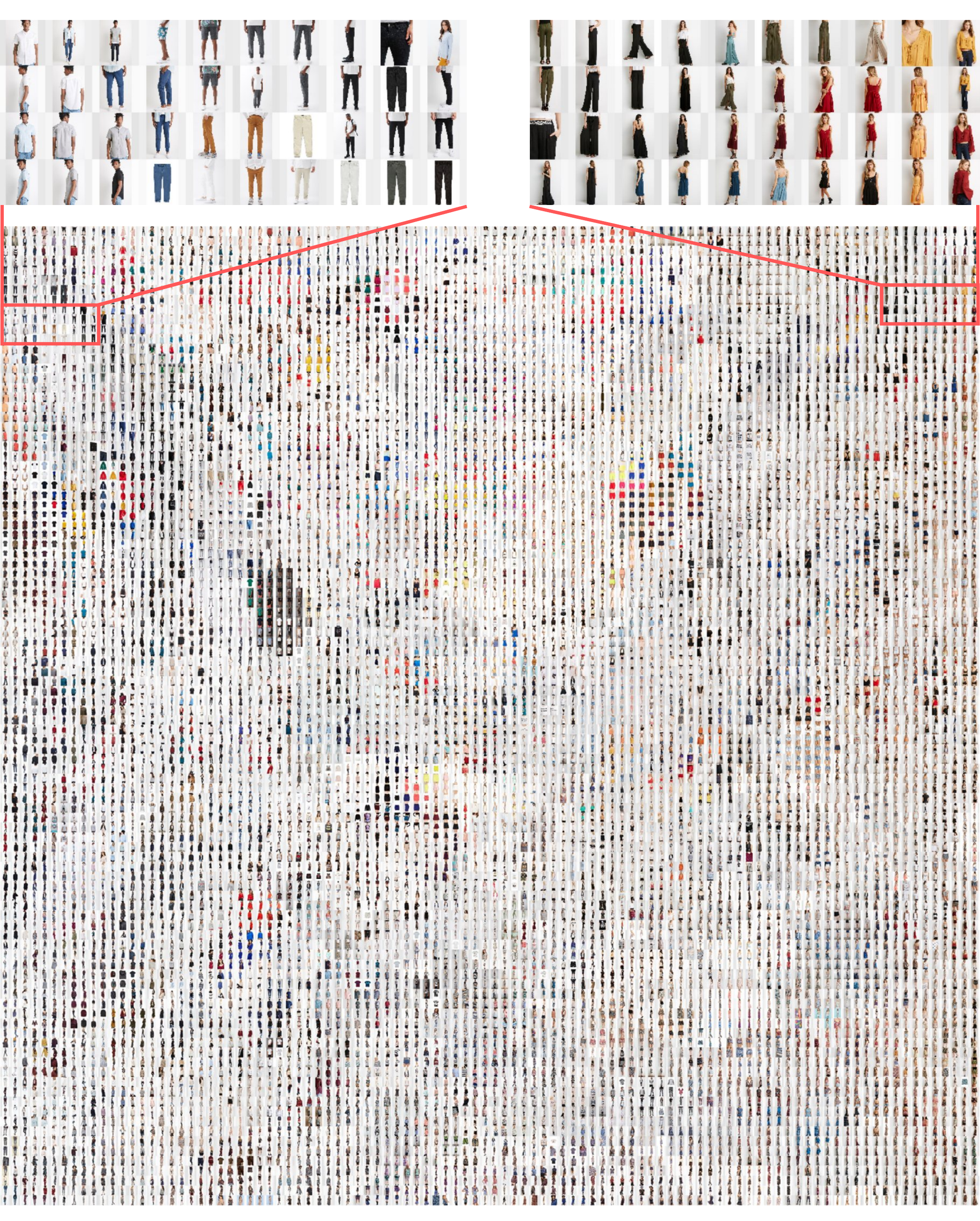}
\caption{t-SNE visualization of our embedding space learned on the test split of In-shop dataset in a grid.
} 
\label{fig:Inshop_tsne}
\end{figure*}

Finally, Figures~\ref{fig:cub_tsne}, \ref{fig:cars_tsne}, \ref{fig:SOP_tsne} and \ref{fig:Inshop_tsne} show t-SNE visualizations of the embedding spaces learned by our loss on the test splits of the four benchmark datasets. These 2D visualizations are generated by mapping each image onto a location of a square grid using Jonker-Volgenant algorithm. The results demonstrate that all data points in the embedding space have relevant nearest neighbors, which suggest that our model learns a semantic similarity that can be generalized even in the test set.

\clearpage


\end{document}